\documentclass[sort&compress,number,preprint,review,12pt]{elsarticle}

\usepackage{nomencl} 
\usepackage{amssymb}
\usepackage[cmex10]{amsmath}
\usepackage{bm}    
\usepackage{graphicx}
\usepackage{float}
\usepackage{subfig}
\usepackage{algorithm}
\usepackage{algorithmic}
\usepackage{url}
\usepackage{rotating}
\usepackage{multirow}
\usepackage{array}
\usepackage{etoolbox}
\usepackage{setspace}

\let\oldtabular\tabular
\let\endoldtabular\endtabular
\renewenvironment{tabular}
   {\bgroup\singlespacing\oldtabular}%
   {\endoldtabular\egroup}

\usepackage{booktabs}

\usepackage{url}
\makeatletter
\def\url@leostyle{%
  \@ifundefined{selectfont}{\def\UrlFont{\sf}}{\def\UrlFont{\small\ttfamily}}}
\makeatother
\begin{document}

\begin{frontmatter}



\title{Supervised Dictionary Learning and Sparse Representation-A Review}


\author[SB,UoT]{Mehrdad~J.~Gangeh\corref{cor}}
\ead{mehrdad.gangeh@utoronto.ca}
\author[UW]{Ahmed~K.~Farahat}
\ead{afarahat@pami.uwaterloo.ca}
\author[UWSTAT]{Ali~Ghodsi}
\ead{aghodsib@uwaterloo.ca}
\author[UW]{Mohamed~S.~Kamel}
\ead{mkamel@pami.uwaterloo.ca}

\cortext[cor]{Corresponding author}

\address[SB]{Departments of Medical Biophysics, and Radiation Oncology, University of Toronto, Toronto, ON M5G 2M9, Canada}
\address[UoT]{Departments of Radiation Oncology, and Imaging Research - Physical Sciences, Sunnybrook Health Sciences Center, Toronto, ON M4N 3M5, Canada}
\address[UW]{Center for Pattern Analysis and Machine Intelligence, Department of Electrical and Computer Engineering, University of Waterloo, 200 University Avenue West, Waterloo, ON N2L 3G1, Canada}
\address[UWSTAT]{Department of Statistics and Actuarial Science, University of Waterloo, 200 University Avenue West, Waterloo, ON N2L 3G1, Canada}

\begin{abstract}
Dictionary learning and sparse representation (DLSR) is a recent and successful mathematical model for data representation that achieves state-of-the-art performance in various fields such as pattern recognition, machine learning, computer vision, and medical imaging. The original formulation for DLSR is based on the minimization of the reconstruction error between the original signal and its sparse representation in the space of the learned dictionary. Although this formulation is optimal for solving problems such as denoising, inpainting, and coding, it may not lead to optimal solution in classification tasks, where the ultimate goal is to make the learned dictionary and corresponding sparse representation as discriminative as possible. This motivated the emergence of a new category of techniques, which is appropriately called supervised dictionary learning and sparse representation (S-DLSR), leading to more optimal dictionary and sparse representation in classification tasks. Despite many research efforts for S-DLSR, the literature lacks a comprehensive view of these techniques, their connections, advantages and shortcomings. In this paper, we address this gap and provide a review of the recently proposed algorithms for S-DLSR. We first present a taxonomy of these algorithms into six categories based on the approach taken to include label information into the learning of the dictionary and/or sparse representation. For each category, we draw connections between the algorithms in this category and present a unified framework for them. We then provide guidelines for applied researchers on how to represent and learn the building blocks of an S-DLSR solution based on the problem at hand. This review provides a broad, yet deep, view of the state-of-the-art methods for S-DLSR and allows for the advancement of research and development in this emerging area of research.
\end{abstract}


\end{frontmatter}


\section{Introduction}\label{sec:introduction}
%
%
%
%
There are many mathematical models to describe data with varying degrees of success, among which dictionary learning and sparse representation (DLSR\nomenclature{DLSR}{Dictionary Learning and Sparse Representation}) have attracted the interest of many researchers in various fields. Dictionary learning and sparse representation are two closely-related topics that have roots in the decomposition of signals to some \emph{predefined} basis, such as the Fourier transform. Representation of signals using predefined basis is based on the assumption that these basis are sufficiently general to represent any kind of signal. However, recent research shows that learning the basis\footnote{Here, the term basis is loosely used as the dictionary can be overcomplete, i.e., the number of dictionary elements can be larger than the dimensionality of the data, and its atoms are not necessarily orthogonal and can be linearly dependent.} from data, instead of using off-the-shelf ones, leads to state-of-the-art results in many applications such as audio processing~\cite{DL:Grosse07}, data representation and column selection~\cite{DL:Elhamifar12,CS:Farahat13}, emotion recognition~\cite{Gangeh:TASLP14}, face recognition~\cite{DL:Zhong07,DL:Wright09,DL:MengYang11}, image compression~\cite{DL:Bryt08}, denoising~\cite{DL:Elad06}, and inpainting~\cite{DL:Mairal08}, image super-resolution~\cite{DL:Yang_Jianchao10}, medical imaging~\cite{Gangeh:MICCAI10,Gangeh:BookChapter11,Gangeh:ISBI13}, motion and data segmentation\cite{DL:Rao08,DL:Elhamifar09}, signal classification~\cite{DL:Raina07,DL:Bradley09,DL:JianchaoYang09}, and texture analysis~\cite{texton:varma05,Gangeh:ICIAR11,DL:Xie10,texton:varma09}. In fact, what makes DLSR distinct from the representation using predefined basis is: first, the basis are learned from the data, and second, only a few components in the dictionary are needed to represent the data (sparse representation). This latter attribute can also be seen in the decomposition of signals using some predefined basis such as wavelets~\cite{book:Mallat3rd}.

Although methods for dictionary learning and sparse representation gained popularity in many domains, their performance is sub-optimal in classification tasks, as they do not exploit the label information in the learning of the dictionary atoms and the coefficients of the sparse approximation. This motivates the emergence of a new category of techniques that utilize label information in computing either dictionary, coefficients, or both. This branch of DLSR is called supervised dictionary learning and sparse representation (S-DLSR), and methods for S-DLSR have shown superior performance in a variety of supervised learning tasks~\cite{DL:Mairal08a,DL:Mairal12,Gangeh:TSP13}.

With the several attempts for learning the dictionary and coefficients in a supervised manner, the literature lacks a comprehensive view of these methods and their connections. In this paper, we present a review of the state-of-the-art techniques in S-DLSR, draw connections between methods, and provide a practical guide for applied researchers in this field on how to design an S-DLSR algorithm. In specific, the contributions of this paper are summarized as follows.
\begin{enumerate}
  \item The paper proposes a taxonomy of S-DLSR methods into six categories based on how the label information is included into the learning of the dictionary and/or sparse coefficients. This taxonomy allows the reader to understand the landscape of existing methods and how they relate to each other.
  \item For the major categories, the paper provides a unified mathematical framework for representing the methods in this category.
  \item The paper discusses the advantages and shortcomings of the methods in each category and the applications where the usage of these methods is preferred.
  \item The paper summarizes the state-of-the-art S-DLSR methods based on their building blocks (i.e., dictionary, sparse coefficients, and the classifier parameters) from the learning and representation perspective and provides guidelines to the applied researchers in the field on how to design these building blocks based on the application at hand.
\end{enumerate}

The comprehensive view of S-DLSR methods presented in this paper will facilitate further contributions in this interesting and useful area of research, and allows the applied researchers to build efficient and effective solutions for different applications.

The rest of the paper is organized as follows.
Section~\ref{sec:background} provides the background for and the related topics to dictionary learning and sparse representation. Particularly, in Subsection~\ref{sec:UnsupervisedDL}, we present the classical formulation of DLSR as an unsupervised dictionary learning approach, which is mainly optimized for the applications such as coding and denoising where the reconstruction of the original signals as accurate as possible is the main concern. In Section~\ref{sec:litReviewSDL}, the main supervised dictionary learning and sparse representation (S-DLSR) methods proposed in the literature are reviewed and categorized depending on how the category information is included into the learning of the dictionary and/or sparse coefficients. Section~\ref{sec:Summary} provides a summary for the S-DLSR methods and how to build them based on three building blocks, i.e., the dictionary learning, sparse representation, and learning the classifier model. Section~\ref{sec:Conclusion} concludes the paper.

\section{Background}\label{sec:background}

\subsection{Related Topics}
\begin{figure*}[!tb]
  \centering
  \includegraphics[width=.8\textwidth]{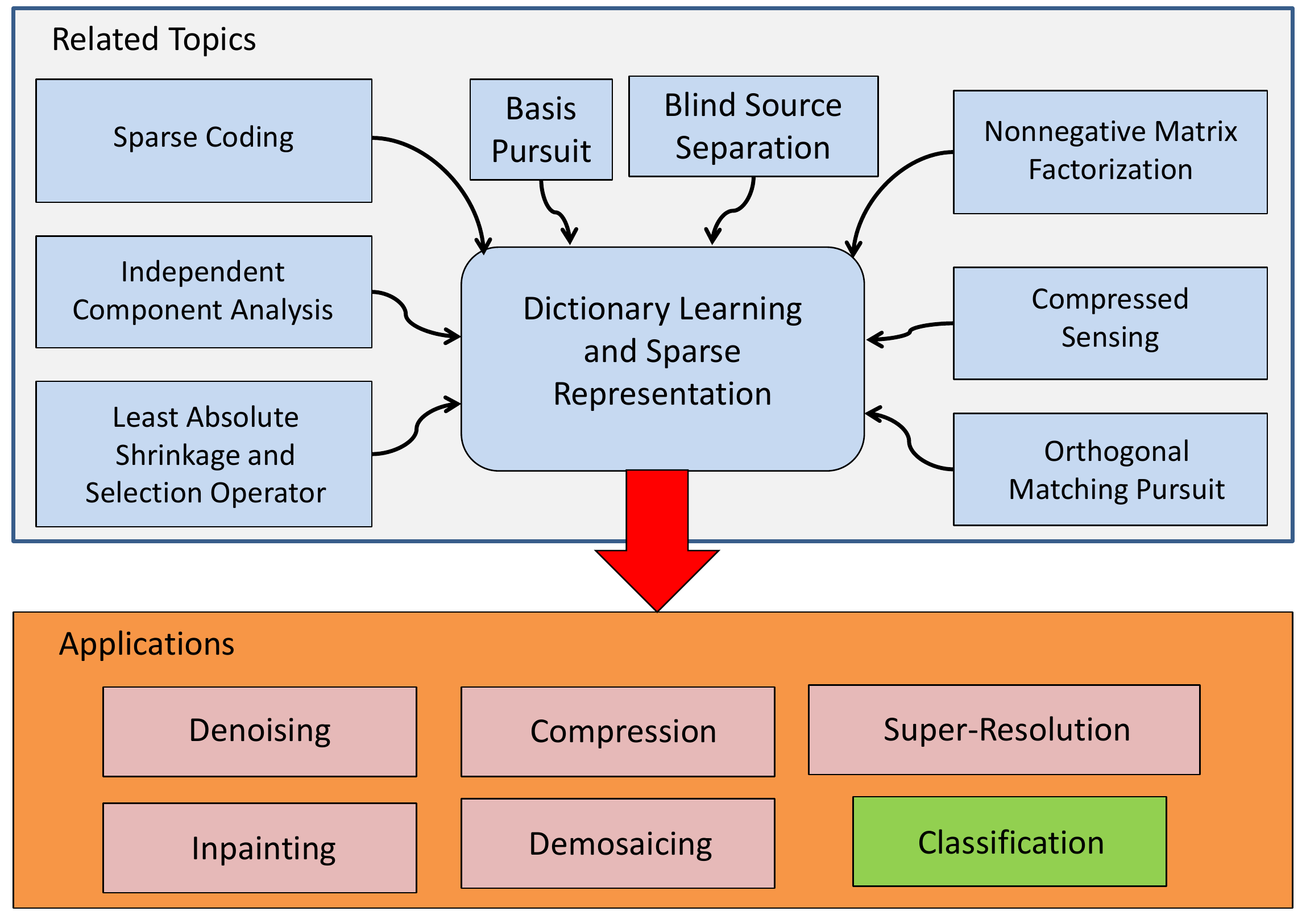}
  \caption{Topics related to and the applications of dictionary learning and sparse representation.}
  \label{fig:relatedTopics}
\end{figure*}

The concept of dictionary learning and sparse representation originated in different communities attempting to solve different problems, which are given different names. Some of these problems are: sparse coding (SC\nomenclature{SC}{Sparse Coding}), which was originated by neurologists as a model for simple cells in mammalian primary visual cortex~\cite{DL:Olshausen96,DL:Olshausen97}; independent component analysis (ICA\nomenclature{ICA}{Independent Component Analysis}), which was developed by researchers in signal processing to estimate the underlying hidden components of multivariate statistical data (refer to~\cite{book:HyvarinenICA,book:StoneICA} for a review of ICA); least absolute shrinkage and selection operator (\emph{lasso}\nomenclature{LASSO}{Least Absolute Shrinkage and Selection Operator}), which was used by statisticians to find linear regression models when there are many more predictors than samples, and some constraints have to be considered to fit the model. In the \emph{lasso}, one of the constraints introduced by Tibshirani was the $\ell_1$ norm that led to sparse coefficients in the linear regression model~\cite{DL:Tibshirani96}. Another technique that also leads to DLSR is nonnegative matrix factorization (NNMF\nomenclature{NNMF}{Nonnegative Matrix Factorization}), which aims at decomposing a matrix to two nonnegative matrices, one of which can be considered to be the dictionary, and the other the coefficients~\cite{DL:Lee99}. In NNMF, usually both the dictionary and coefficients are sparse~\cite{DL:Biggs08,DL:Hoyer04}. This list is not complete, and there are variants for each of the above techniques, such as blind source separation (BSS\nomenclature{BSS}{Blind Source Separation})~\cite{DL:Jutten91}, compressed sensing~\cite{DL:Donoho06}, basis pursuit (BP\nomenclature{BP}{Basis Pursuit})~\cite{DL:Chen98}, and orthogonal matching pursuit (OMP\nomenclature{OMP}{Orthogonal Matching Pursuit})~\cite{DL:Pati93,DL:Mallat93}. The reader is referred to~\cite{DL:Wright10,DL:Bruckstein09,book:Elad,DL:Rubinstein10} for some reviews on these techniques. Figure~\ref{fig:relatedTopics} summarizes the topics related to and the applications of dictionary learning and sparse representation.

The main results of all these research efforts is that a class of signals with sparse nature, such as images of natural scenes, can be represented using some \emph{primitive elements} that form a dictionary, and each signal in this class can be represented by using only a few elements in the dictionary, i.e., by a sparse representation. In fact, there are at least two ways in the literature to exploit sparsity~\cite{DL:Mairal08a}: first, using a linear/nonlinear combination of some predefined basis, e.g., wavelets~\cite{book:Mallat3rd}; second, using primitive elements in a learned dictionary, such as the techniques employed in SC or ICA. This latter approach is the focus of this paper.

\subsection{Taxonomy of DLSR Methods}
One may categorize the various dictionary learning with sparse representation approaches proposed in the literature in different ways: one where the dictionary consists of predefined or learned basis as stated above, and the other based on the model used to learn the dictionary and coefficients. These models can be \emph{generative} as used in the original formulation of SC~\cite{DL:Olshausen96}, ICA~\cite{book:HyvarinenICA}, and NNMF~\cite{DL:Lee99}; \emph{reconstructive} as in the \emph{lasso}~\cite{DL:Tibshirani96}; or \emph{discriminative} such as SDL-D\nomenclature{SDL-D}{Supervised Dictionary Learning-Discriminative} (supervised dictionary learning-discriminative) in~\cite{DL:Mairal08a}. The two former approaches do not consider the class labels in building the dictionary, while the discriminative one does. In other words, dictionary learning can be performed unsupervised or supervised, with the difference that in the latter, the class labels in the training set are used to build a more discriminative dictionary for the particular classification task at hand.

\subsection{Unsupervised Dictionary Learning}\label{sec:UnsupervisedDL}

\begin{table*}[!t]
  \tiny
  \caption{The list of notations and their definitions in this paper.}
  \centering
  \label{tab:notations}
    \begin{tabular}{ | c | p{4cm} || c | p{4cm} |}
    \hline
    Notation & Definition & Notation & Definition \\ \hline
    $\mathbf{X}$ & a finite set of data samples & $\mathbf{X}_i$ & the group of data samples in class $i$ \\ \hline
    $\mathbf{x}_i$ & the $i^{\textup{th}}$ data sample & $\mathbf{x}_{ij}$ & the $j^{\textup{th}}$ data sample in class $i$\\ \hline
    $\mathbf{x}_i^j$ & a constituent of signal $\mathbf{x}_i$ & $\mathbf{x}_{\textup{ts}}$ & a test data sample   \\ \hline
    $\mathcal{X}$ & a random variable representing the data samples & $\mathbf{D}$ & dictionary \\ \hline
    $\mathbf{D}_i$ & the subdictionary learned on class $i$ & $\mathbf{d}_i$ & the $i^{\textup{th}}$ column of $\mathbf{D}$ \\ \hline
    $\mathbf{d}_{ij}$ & the $j^{\textup{th}}$ dictionary atom in subdictionary learned in class $i$ & $\mathcal{D}$ & a random variable representing the dictionary atoms  \\ \hline
    $\mathbf{A}$ & sparse coefficients & $\mathbf{A}_i$ & part of sparse coefficients corresponding to class $i$  \\ \hline
    $\mathbf{A}_i^j$ & part of sparse coefficients that represent class $i$ over $\mathbf{D}_j$ & $\bm{\alpha}_i$ & the $i^{\textup{th}}$ column of $\mathbf{A}$ \\ \hline
    $\bm{\alpha}_i^j$ & the sparse coefficient corresponding to signal constituent $\mathbf{x}_i^j$ & $L$, $l$ & loss function \\ \hline
    $\mathbf{Y}$ & class labels   & $\mathcal{Y}$ & a random variable representing class labels \\ \hline
    $\mathbf{h}$ & a histogram  & $\mathcal{H}$ & a random variable representing histograms   \\ \hline
    $\mathbf{H}$ & centering matrix & $\mathbf{I}$ & identity matrix  \\ \hline
    $\mathbf{K}$ & kernel on data & $\mathbf{L}$ & kernel on labels  \\ \hline
    $\mathbf{W}$ & classifier parameters to be learned  & $\mathbf{U}$ & a transformation/projection to be computed \\ \hline
    $\mathbf{Q}$ & optimal discriminative sparse codes & $\mathcal{Q}$ & an incoherence term \\ \hline
    $S_{\textup{B}}$ & between-class covariance matrix  & $S_{\textup{W}}$ & within-class covariance matrix \\ \hline
    $S_{\beta}$ & A sigmoid function with the slope of $\beta$ & $S_i$ & the $i^{\textup{th}}$ cluster \\ \hline
    $P(.,.)$ & joint probability  & $I(.,.)$ & mutual information shared by two random variables  \\ \hline
    $R(.)$ & The ratio of intra- to inter-class reconstruction error  & $\mathcal{C}(.)$ & logistic regression function  \\ \hline
    $\delta_i$ & a characteristic function that selects the coefficients associated with class i & $r_i(.)$ & the residual error between a data sample and its reconstructed version\\ \hline
    $\psi$ & a generic sparsity inducing function & $\lambda$, $\lambda_0$, $\lambda_1$, $\lambda_2$, $\eta$, $\gamma$ & regularization parameters \\ \hline
    $\|.\|_{\textup{F}}$ & Frobenius norm  & $\|.\|_1$ & $\ell_1$ norm \\ \hline
    $\mathbf{e}$ & a vector of all ones  & $\textup{tr}(.)$ & trace operator   \\ \hline
    $n$ & the number of data samples & $m$ & the number of data samples in a class \\ \hline
    $p$ & the dimensionality of data & $c$ & the number of classes   \\ \hline
    $k$ & the number of dictionary atoms & $k_i$ & the number of dictionary atoms in class $i$   \\ \hline
    \end{tabular}
\end{table*}

Considering a finite training set of signals\footnote{For the convenience of readers, the list of main notations in this review paper is provided in Table~\ref{tab:notations}.} $ \mathbf{X} = [\mathbf{x}_1,\mathbf{x}_2,...,\mathbf{x}_n]\in\mathbb{R}^{p \times n} $, where $p$ is the dimensionality and $n$ is the number of data samples, according to classical dictionary learning and sparse representation (DLSR) techniques (refer to~\cite{DL:Wright10,DL:Bruckstein09,book:Elad} for a recent review on this topic), these signals can be represented by a linear decomposition over a few dictionary atoms by minimizing a loss function as given below
\begin{equation}\label{eq:loss1}
L(\mathbf{X},\mathbf{D},\mathbf{A})=\sum_{i=1}^{n}l(\mathbf{x}_i,\mathbf{D},\mathbf{A}),
\end{equation}
where $L$ and $l$ are the overall and per data sample loss functions, respectively, $\mathbf{D}\in\mathbb{R}^{p \times k}$ is the dictionary of $k$ atoms, and $\mathbf{A}\in\mathbb{R}^{k \times n}$ are the coefficients.

The loss function can be defined in various ways based on the application at hand. However, what is common in DLSR literature is to define the loss function $L$ as the reconstruction error in a mean-squared sense, with a sparsity-inducing function $\psi$ as a regularization penalty to ensure the sparsity of coefficients. Hence, (\ref{eq:loss1}) can be written as
\begin{equation}\label{eq:loss2}
L(\mathbf{X},\mathbf{D},\mathbf{A})=\min_{\mathbf{D},\mathbf{A}}\frac{1}{2}\|\mathbf{X}-\mathbf{D}\mathbf{A}\|_{\textup{F}}^{2}+\lambda\psi(\mathbf{A}),
\end{equation}
where subscript $\textup{F}$ indicates the Frobenius norm\footnote{The Frobenius norm of a matrix $\mathbf{X}$ is defined as $\|\mathbf{X}\|_{\textup{F}}=\sqrt{\sum_{i,j}(x_{i,j}^2)}$.} and $\lambda$ is the regularization parameter that affects the number of nonzero coefficients.

An intuitive measure of sparsity is $\ell_0$ norm\footnote{The $\ell_0$ norm of a vector $\mathbf{x}$ is defined as $\|\mathbf{x}\|_{0}=\#\{i:x_i\neq0\}$.}, which indicates the number of nonzero elements in a vector. However, the optimization problem obtained from replacing sparsity-inducing function $\psi$ in (\ref{eq:loss2}) with $\ell_0$ is non-convex and NP-hard (refer to~\cite{DL:Bruckstein09} for a recent comprehensive discussion on this issue). Two main categories of approximate solutions have been proposed to overcome this problem: the first is based on greedy algorithms, such as the well-known orthogonal matching pursuit (OMP)~\cite{DL:Pati93,DL:Mallat93,DL:Bruckstein09}; the second works by approximating a highly discontinuous $\ell_0$ norm by a continuous function such as the $\ell_1$ norm. This leads to an approach which is  widely known in the literature as \emph{lasso}~\cite{DL:Tibshirani96} or \emph{basis pursuit} (BP)~\cite{DL:Chen98}, and~(\ref{eq:loss2}) converts to
\begin{equation}\label{eq:lasso}
L(\mathbf{X},\mathbf{D},\mathbf{A})=\min_{\mathbf{D},\mathbf{A}}\sum_{i=1}^{n}\begin{pmatrix}\frac{1}{2}\|\mathbf{x}_i-\mathbf{D}\bm{\alpha}_i\|_{2}^{2}+\lambda\|\bm{\alpha}_i\|_{1}\end{pmatrix},
\end{equation}
where $\mathbf{x}_i$ is the $i^{\textup{th}}$ training sample and $\bm{\alpha}_i$ is the $i^{\textup{th}}$ column of $\mathbf{A}$.

The reconstructive formulation given in~(\ref{eq:lasso}) is non-convex when both the dictionary $\mathbf{D}$ and coefficients $\mathbf{A}$ are unknown. However, this optimization problem is convex if it is solved iteratively and alternately on these two unknowns. Several fast algorithms have recently been proposed for this purpose, such as K-SVD~\cite{DL:Aharon06}, online learning~\cite{DL:Mairal09,DL:Mairal10}, and cyclic coordinate descent~\cite{DL:Friedman10}. 

In~(\ref{eq:lasso}), the main optimization goal for the computation of the dictionary and sparse coefficients is minimizing the reconstruction error in the mean-squared sense. While this works well in applications where the primary goal is to reconstruct signals as accurately as possible, such as in denoising, image inpainting, and coding, it is not the ultimate goal in classification tasks as discriminating signals is more important here~\cite{DL:Huang07}. Recently, there have been several attempts to include category information in computing either dictionary, coefficients, or both. This branch of DLSR is called supervised dictionary learning and sparse representation (S-DLSR). In the following section, an overview of proposed S-DLSR approaches in the literature will be provided.

\section{Taxonomy of Supervised Dictionary Learning and Sparse Representation Techniques}\label{sec:litReviewSDL}

In this section, the proposed \emph{supervised} dictionary learning and sparse representation (S-DLSR) approaches in the literature are categorized into six different groups, depending on how the class labels are included into the learning of the dictionary and/or sparse coefficients. These six categories are: 1)~learning one dictionary per class, 2)~unsupervised dictionary learning followed by supervised pruning, 3)~joint dictionary and classifier learning, 4)~embedding class labels into the learning of dictionary, 5)~embedding class labels into the learning of sparse coefficients, and 6)~learning a histogram of dictionary elements over signal constituents. We admit that the taxonomy proposed in this section is not unique and could be done differently. Also, it is worthwhile to mention that while the first five categories perform S-DLSR on whole signal, the last category performs it on signal constituents. In the rest of this section, the six categories are described and their advantages and disadvantages are discussed in details.

\subsection{Learning One Dictionary per Class}\label{subsec:perClass}

The first and simplest approach to include category information in DLSR is computing one dictionary per class, i.e., using the training samples in each class to compute part of the dictionary, and then composing all these partial dictionaries into one. In providing the mathematical formulation for all the approaches in this category of S-DLSR, it is always assumed that the training samples are grouped based on the classes they belong to such that $\mathbf{X} = [\mathbf{X}_1,\mathbf{X}_2,...,\mathbf{X}_c]\in\mathbb{R}^{p \times n}$, where $c$ is the number of classes and $\mathbf{X}_i=[\mathbf{x}_{i1},\mathbf{x}_{i2},...,\mathbf{x}_{im}]\in\mathbb{R}^{p \times m}$ is the group of $m$ training samples in class $i$. Similarly, the dictionary $\mathbf{D}$ is described as $\mathbf{D} = [\mathbf{D}_1,\mathbf{D}_2,...,\mathbf{D}_c]\in\mathbb{R}^{p \times k}$, where $\mathbf{D}_i=[\mathbf{d}_{i1},\mathbf{d}_{i2},...,\mathbf{d}_{ik_{i}}]\in\mathbb{R}^{p \times k_i}$ is the subdictionary of $k_i$ atoms in class $i$.

Among the methods in this category, the most common ones are: 1)~supervised $k$-means, 2)~sparse representation-based classification (SRC), 3)~metaface, and 4)~dictionary learning with structured incoherence (DLSI). These methods are described in the rest of this subsection.

\subsubsection{Supervised k-means}

Perhaps the earliest work in this direction is the one based on the so-called texton-based approach~\cite{texton:Julesz81,texton:Leung01,texton:Cula04,texton:Schmid04,texton:varma05,texton:varma09}. The texton-based approach, can be considered as a dictionary learning approach particularly tailored for texture analysis. In this approach, textons, which are computed using the \emph{k}-means clustering algorithm over patches extracted from texture images, play the role of dictionary atoms. Although in a texton-based approach, the texture images are usually modeled with a histogram of textons, i.e., using a model of signal constituents, and hence, the approach falls mainly into the category of S-DLSR explained in Subsection~\ref{subsec:histogram}, the idea of using \emph{k}-means and the computed cluster centers as the dictionary elements can still be considered here as an S-DLSR approach that computes one dictionary per class. Therefore, a specific name is suggested for this technique, i.e., supervised \emph{k}-means, to differentiate it from the texton-based approach. In supervised \emph{k}-means, the \emph{k}-means algorithm is applied to the training samples in each class, and the \emph{k} cluster centers computed are considered to be the dictionary for this class. These partial dictionaries are eventually composed into one dictionary.

In the mathematical framework, each subdictionary $\mathbf{D}_i=[\mathbf{d}_{i1},\mathbf{d}_{i2},...,\mathbf{d}_{ik_{i}}]\in \mathbb{R}^{p \times k_{i}}$ can be computed using the training samples in class $i$, i.e., using $\mathbf{X}_i=[\mathbf{x}_{i1},\mathbf{x}_{i2},...,\mathbf{x}_{im}]\in\mathbb{R}^{p \times m}$ and the optimization problem
\begin{equation}\label{eq:supk-means}
\underset{\mathbf{D}_i}{\text{arg min}}\sum_{l=1}^{k_i}\sum_{\mathbf{x}_{ij}\in S_l}\left \| \mathbf{x}_{ij}-\mathbf{d}_{il} \right \|
\end{equation}
where $S=\left \{ S_1,S_2,...,S_{k_{i}} \right \}$ are $k_i$ clusters that partition data samples $\mathbf{X}_i$ in class $i$. Usually, $k_i$, the number of dictionary atoms computed per class, is the same over all classes. By composing all $\mathbf{D}_i$ into one dictionary such that $\mathbf{D} = [\mathbf{D}_1,\mathbf{D}_2,...,\mathbf{D}_c]\in\mathbb{R}^{p \times k}$, where $k=k_i\cdot c$, the whole dictionary is obtained.

One can explain why it might be expected that a supervised \emph{k}-means performs better than an unsupervised one by understanding how \emph{k}-means compute the cluster centers: it essentially computes the cluster centers by taking the mean of the points. Hence, if \emph{k}-means was applied to the data points across classes, the resultant cluster centers might not correspond to the data points in any of the classes, and consequently the resultant cluster centers would not be identified uniquely with individual classes. In other words, the cluster centers computed using \emph{k}-means across classes would not be representing data samples in a class properly. Thus, in classification tasks, it will be beneficial, particularly at small dictionary sizes, to use \emph{k}-means for the data points in one class at a time~\cite[Table II]{Gangeh:TSP13}.

\subsubsection{Sparse representation-based classification (SRC\nomenclature{SRC}{Sparse Representation-based Classification})}

In their seminal work, Wright \emph{et al.}~\cite{DL:Wright09} proposed to use the training samples as the dictionary in a technique called sparse representation-based classification (SRC). The approach was proposed in the application of face recognition and effectively falls into the same category as training one dictionary per class. However, no actual training is performed here, and the whole training samples are used directly in the dictionary.

To describe SRC more formally, suppose that $\mathbf{x}_{\textup{ts}}\in\mathbb{R}^{p}$ is a test sample. The SRC algorithm assigns the whole training set $\mathbf{X}$ to the dictionary $\mathbf{D}$ such that $\mathbf{D_i}=\mathbf{X}_i$ for class $i$, and computes the sparse coefficients $\bm{\alpha}$ for the test sample $\mathbf{x}_{\textup{ts}}$ using the \emph{lasso} given in (\ref{eq:lasso}) as follows
\begin{equation}\label{eq:SRCLasso}
\min_{\bm{\alpha}}\frac{1}{2}\|\mathbf{x}_{\textup{ts}}-\mathbf{X}\bm{\alpha}\|_{2}^{2}+\lambda\|\bm{\alpha}\|_{1}.
\end{equation}

In the next step, the residual error is computed for the reconstruction of the test sample using the training samples of each class and their corresponding sparse coefficients
\begin{equation}\label{eq:SRCResidue}
r_i(\mathbf{x}_{\textup{ts}})=\|\mathbf{x}_{\textup{ts}}-\mathbf{X}\delta_i(\bm{\alpha})\|_{2}^{2},
\end{equation}
where $\delta_i$ is a characteristic function that selects the coefficients associated with class $i$. This residual error is found for each class separately, and then the class label of the given test sample is assigned according to
\begin{equation}\label{eq:SRCResidue1}
label(\mathbf{x}_{\textup{ts}})=\underset{i}{\text{arg min}}\;\;r_i(\mathbf{x}_{\textup{ts}}).
\end{equation}

For a low to moderate training set size, this approach is computationally very efficient as there is no overhead for the learning of the dictionary. Moreover, using minimum residual error for the purpose of classification of an unseen test sample is easily interpretable as the class of the subdictionary leading to minimum residual error can be inspected and assigned as the class label of the test sample. The main disadvantage of this method, however, is that using the training samples as the dictionary in this approach may result in a very large and possibly inefficient dictionary, due to the noisy training instances. This is particularly the case in applications with large training set sizes.

\subsubsection{Metaface}

To obtain a smaller dictionary, Yang \emph{et al.} proposed an approach called \emph{metaface}, which learns a smaller dictionary for each class and then composes them into one dictionary~\cite{DL:MengYang10}. Metaface was originally proposed for the application of face recognition, but it is general and can be used in any application. In this approach, each subdictionary $\mathbf{D}_i$ is computed using the training samples $\mathbf{X}_i$ in class $i$ using the formulation given in~(\ref{eq:lasso}) as follows\footnote{In this paper, whenever $\ell_1$ norm is used over a matrix, it is meant that $\ell_1$ norms over each column of the matrix are summed such as what is used in~(\ref{eq:lasso}). Hence the correct form for~(\ref{eq:metaface}) is: $\min_{\mathbf{D}_i,\mathbf{A}_i}\sum_{j=1}^{m}\begin{pmatrix}\frac{1}{2}\|\mathbf{x}_{ij}-\mathbf{D}\mathbf{A}_{ij}\|_{2}^{2}+\lambda\|\mathbf{A}_{ij}\|_{1}\end{pmatrix}$. However, similar forms as in~(\ref{eq:metaface}) are loosely used for $\ell_1$ norm in the rest of this paper to avoid too long and complex formulations and to focus more on the concept.}
\begin{equation}\label{eq:metaface}
\min_{\mathbf{D}_i,\mathbf{A}_i}\frac{1}{2}\|\mathbf{X}_i-\mathbf{D}_i\mathbf{A}_i\|_{\textup{F}}^{2}+\lambda\|\mathbf{A}_i\|_{1}.
\end{equation}
where $\mathbf{A}_i\in\mathbb{R}^{k_i\times m}$ is the matrix of sparse coefficients representing $\mathbf{X}_i$.

Since this optimization problem is non-convex when both dictionary and coefficients are unknown, it has to be solved iteratively and alternately with one unknown variable considered fixed in each alteration. Computed subdictionaries are eventually composed into one dictionary $\mathbf{D} = [\mathbf{D}_1,\mathbf{D}_2,...,\mathbf{D}_c]\in\mathbb{R}^{p \times k}$. After the computation of the dictionary, the class label of a test sample $\mathbf{x}_{\textup{ts}}$ is computed in the same way as explained in the SRC approach, i.e., by finding the coefficients for this test sample using the computed dictionary instead of the whole training set in~(\ref{eq:SRCLasso}), followed by the computation of the residuals given in~(\ref{eq:SRCResidue}), and assigning the test sample to the class that yields the minimal residue.

Although the metaface approach can potentially reduce the size of the dictionary compared to the SRC method, its major drawback is that the training samples in one class are used for computing the atoms in the corresponding subdictionary, irrespective of the training samples from other classes. This means that if the training samples across classes have some common properties, these shared properties cannot be learned in common in the dictionary.

\subsubsection{Dictionary learning with structured incoherence (DLSI\nomenclature{DLSI}{Dictionary Learning with Structured Incoherence})}

Ramirez \emph{et al.} proposed to overcome the aforementioned problem with the metaface approach by including an incoherence term in~(\ref{eq:lasso}) to encourage independency of dictionaries from different classes, while still allowing for different classes to share features~\cite{DL:Ramirez:10}.

To enable sharing features among the data points in different classes for learning the dictionary, instead of learning each $\mathbf{D}_i$ independently and unaware of data points in other classes, a coherence term is added to the \emph{lasso} as described by the formulation below
\begin{equation}\label{eq:DLSI}
\underset{\mathbf{D},\mathbf{A}}{\text{min}}\sum_{i=1}^{c}\begin{Bmatrix} \left \|  \mathbf{X}_i-\mathbf{D}_i\mathbf{A}_i\right \|_{\textup{F}}^2  +\lambda \left \| \mathbf{A}_i \right \|_1 \end{Bmatrix} +\eta \sum_{i\neq j}\left \| \mathbf{D}_i^{\top}\mathbf{D}_j  \right \|_{\textup{F}}^2,
\end{equation}
where the last term is an incoherence term $\mathcal{Q}(\mathbf{D}_i,\mathbf{D}_j)$, which has been proposed in~\cite{DL:Ramirez:10} to be the inner product between the two subdictionaries\footnote{Please note that the last term in~(\ref{eq:DLSI}) is an inner product and hence, a measure of similarity/coherence. However, since this term has been minimized in the optimization problem, it is called incoherence term by the authors in the original paper, which is also adopted here.} $\mathbf{D}_i$ and $\mathbf{D}_j$, but it could be defined differently as long as it includes some measure of (dis)similarity/(in)coherence. In fact, the incoherence term in~(\ref{eq:DLSI}) discourages the similarity among the subdictionaries learned across different classes. After finding the dictionary, the classification of a test sample is performed the same way as with the SRC.

\paragraph*{Discussion} The advantage of the S-DLSR methods in this category is mainly the ease of the computation of the dictionary. In case of the SRC method, no learning is needed for the dictionary as the dictionary is the same as the training samples. However, the main drawback of all the approaches in this category is that they may lead to a very large dictionary, as the size of the composed dictionary grows linearly with the number of classes. An example is in face recognition where there are many classes. For example, in Extended Yale B database~\cite{Face:Georghiades01}, there are 38 classes and learning even 10 atoms per class (in SRC, all data instances in the training set are included in the dictionary) can easily lead to a large dictionary.

\subsection{Unsupervised Dictionary Learning Followed by Supervised Pruning}\label{subsec:Pruning}

The second category of S-DLSR approaches learns a very large dictionary unsupervised in the beginning, then merges the atoms in the dictionary by optimizing an objective function that takes into account the category information. the two main methods in this category are: 1)~an approach based on information bottleneck (IB), and 2)~universal visual dictionary (UVD). The details of these methods are as follows.

\subsubsection{Information bottleneck (IB\nomenclature{IB}{Information Bottleneck})}

One major work in the literature in this direction is based on agglomerative information bottleneck (AIB\nomenclature{AIB}{Agglomerative Information Bottleneck}), which iteratively merges two dictionary atoms that cause the smallest decrease in the mutual information between the dictionary atoms and the class labels~\cite{DL:Fulkerson08}. The discriminative power of a dictionary $\mathbf{D}$ is characterized by the AIB as the amount of mutual information $I(\mathcal{D},\mathcal{Y})$ shared by random variables $\mathcal{D}$ (dictionary atoms) and $\mathcal{Y}$ (category information):
\begin{equation}
I(\mathcal{D},\mathcal{Y})=\sum_{d\in \mathcal{D}}\sum_{y\in \mathcal{Y}}P(d,y)log\frac{P(d,y)}{P(d)P(y)}
\end{equation}
where the joint probability $P(d,y)$ is estimated from the data by counting the number of occurrences of dictionary atoms $d$ in each category $y=\{1,...,c\}$. The mutual information $I(d,y)$ is monotonically decreased as the AIB iteratively compresses the dictionary by merging dictionary atoms such that smallest decrease in the mutual information (discriminating power) $I(\mathcal{D},\mathcal{Y})$ occurs. This is continued until a predefined dictionary size is obtained. Although the approach is slow, a solution called ``Fast AIB'' has been proposed in~\cite{DL:Fulkerson08} to make it computationally efficient.

\subsubsection{Universal visual dictionary (UVD\nomenclature{UVD}{Universal Visual Dictionary})}

Another major work is based on merging two dictionary atoms so as to minimize the loss of mutual information between the histogram of dictionary atoms over signal constituents, e.g., image patches, and class labels~\cite{DL:Winn05}. From this point of view, the difference between this approach and the one based on AIB is in the way they measure the discriminative power of the dictionary. In this approach, rather than measuring the discriminative power of the dictionary on individual dictionary atoms, it is measured on the histogram of dictionary atoms $\mathbf{h}$ over signal constituents. Therefore, $I(\mathcal{H},\mathcal{Y})$, where $\mathcal{H}$ is the random variable over the histograms $\mathbf{h}$ is considered in UVD, instead of $I(\mathcal{D},\mathcal{Y})$ used by AIB. However, since the dimensionality of histograms tends to be very high, the estimation of $I(\mathcal{H},\mathcal{Y})$ is only possible with strong assumptions on the histograms. In~\cite{DL:Winn05}, it is assumed that histograms can be modeled using a mixture of Gaussians, with one Gaussian per category. Based on this assumption, in~\cite{DL:Winn05}, category posterior probability $p(y|h)$ is used instead of mutual information $I(\mathcal{H},\mathcal{Y})$ for characterizing the discriminative power of the dictionary. Since this approach works on a histogram of dictionary atoms over signal constituents, it can also be categorized in the sixth category of S-DLSR explained in Subsection~\ref{subsec:histogram}.

\paragraph*{Discussion} One main drawback of this category of S-DLSR is that the reduced dictionary obtained performs, at best, as good as the original one. Since the initial dictionary is learned in an unsupervised manner, even though with its large size, it includes almost all possible atoms that helps to improve the performance of the classification task~\cite{DL:Philbin07,DL:Tuytelaars07,DL:Moosmann06}, the consecutive pruning stage is inefficient in terms of computational load. This might be one of the reasons that this category of S-DLSR has attracted less attention among other S-DLSR approaches in the literature as the efficiency of the method can significantly be improved by finding a discriminative dictionary from the beginning.

\subsection{Joint Dictionary and Classifier Learning}\label{subsec:jointDLClassifier}

The third category of S-DLSR, which is based on several research works published in~\cite{DL:Mairal08a,DL:Mairal08b,DL:QiangZhang10,DL:Pham08,DL:LiuYang08,DL:Mairal12} can be considered a major leap in the field. In this category, the classifier parameters and the dictionary are learned in a joint optimization problem. The main methods in this category are: 1)~supervised dictionary learning-discriminative (SDL-D), 2)~discriminative K-SVD (DK-SVD), 3)~label consistent K-SVD (LCK-SVD), and 4)~Bayesian supervised dictionary learning, which are described in the following subsections.

\subsubsection{Supervised dictionary learning-discriminative (SDL-D)}

Mairal \emph{et al.} were one of the first research teams who proposed a joint optimization problem for learning the dictionary and the classifier parameters~\cite{DL:Mairal08a,DL:Mairal08b,DL:Mairal12}. In~\cite{DL:Mairal08a} they proposed the following formulation
\begin{align}\label{eq:SDL-G}
\min_{\mathbf{D},\mathbf{W},\mathbf{A}}\Bigl(
\sum_{i=1}^{n}\mathcal{C}(y_{i}f(\mathbf{x}_{i},\bm{\alpha}_{i},\mathbf{W})) +\lambda_{0}\left \| \mathbf{x}_{i}-\mathbf{D}\bm{\alpha}_{i} \right \|_{2}^{2}
  \mspace{1mu}
  \notag\\
+ \lambda_{1}\left \| \bm{\alpha}_{i} \right \|_{1}
\Bigr) + \lambda_{2}\left \| \mathbf{W} \right \|_{\textup{F}}^{2},
\end{align}
where $\mathcal{C}(x) = log(1+e^{-x})$ is the logistic loss function, $(y_i\in \{-1,+1\})_{i=1}^{n}$ are binary class labels\footnote{The approach can be easily extended to multiclass problem.}, $f(.)$ is the classifier function, and $\mathbf{W}$ is the associated classifier parameters to be learned. In~(\ref{eq:SDL-G}), $\lambda_{0}$ is the parameter that controls the relative importance of the reconstruction error and the loss function on the classifier, $\lambda_{1}$ is the regularization parameter that controls the level of sparsity of the coefficients, and $\lambda_{2}$ is the regularization parameter to prevent overfitting the classifier. The actual discriminative formulation proposed in~\cite{DL:Mairal08a} is sufficiently more complex than~(\ref{eq:SDL-G}) and its description is not provided here. The optimization problem in~(\ref{eq:SDL-G}), is a non-convex problem and has many parameters to tune, which makes the approach computationally expensive.

\subsubsection{Discriminative K-SVD (DK-SVD\nomenclature{DK-SVD}{Discriminative K-SVD})}

In~\cite{DL:QiangZhang10}, Zhang and Li proposed a technique called discriminative K-SVD (DK-SVD).  DK-SVD truly jointly learns the classifier parameters and dictionary, without alternating between these two steps. This prevents the possibility of the solution to get stuck in some local minima. However, only linear classifiers are considered in DK-SVD, which may lead to poor performance in difficult classification tasks.

To provide the formulation for DK-SVD, one may notice that after learning the dictionary using the \emph{lasso}~(\ref{eq:lasso}), a linear classifier is to be learned on the coefficients $\mathbf{A}$ in the space of learned dictionary. Suppose that $\mathbf{W}\in\mathbb{R}^{c \times k}$ is the matrix of classifier parameters ($c$ is the total number of classes and $k$ is the number of dictionary atoms), and $\mathbf{Y}\in\mathbb{R}^{c \times n}$ includes the class labels ($n$ is the number of training samples) such that each column of $\mathbf{Y}$ is $\mathbf{y}_{i}=\{0,...,1,...,0\}^{\top}$, i.e., there is exactly one nonzero element in each column of $\mathbf{Y}$, whose position indicates the class of the corresponding training sample. The classifier can be learned using least square formulation by minimizing the classifier error in the mean-squared sense using the optimization problem as follows
\begin{equation}
\min_{\mathbf{W}}\frac{1}{2}\|\mathbf{Y}-\mathbf{W}\mathbf{A}\|_{\textup{F}}^{2}.
\end{equation}

This optimization problem can be combined with the \emph{lasso}~(\ref{eq:lasso}) into one optimization problem
\begin{equation}\label{eq:DK-SVDPre}
\min_{\mathbf{D},\mathbf{W},\mathbf{A}}\frac{1}{2}\|\mathbf{X}-\mathbf{D}\mathbf{A}\|_{\textup{F}}^{2}+\frac{\gamma}{2}\|\mathbf{Y}-\mathbf{W}\mathbf{A}\|_{\textup{F}}^{2}+\lambda\|\mathbf{A}\|_{1}.
\end{equation}
To find the dictionary, coefficients, and the classifier, the optimization problem given in~(\ref{eq:DK-SVDPre}) has to be solved iteratively and alternately, with two of these unknowns fixed each time and solving for the third. This makes the solution slow and very likely to get stuck in some local minima. To partially overcome these problems, it is proposed in~\cite{DL:QiangZhang10} to combine the first two terms in~(\ref{eq:DK-SVDPre}) into one term as follows
\begin{equation}\label{eq:DK-SVD}
\min_{\mathbf{D},\mathbf{W},\mathbf{A}}\frac{1}{2}\begin{Vmatrix}\begin{bmatrix}
\mathbf{X} \\ \sqrt{\gamma}\;\mathbf{Y}
\end{bmatrix}-\begin{bmatrix}
\mathbf{D}\\ \sqrt{\gamma}\;\mathbf{W}
\end{bmatrix}\mathbf{A}\end{Vmatrix}_{\textup{F}}^{2} + \lambda\|\mathbf{A}\|_{1}.
\end{equation}
Considering $\begin{bmatrix} \mathbf{X}^{\top} , \sqrt{\gamma}\;\mathbf{Y}^{\top} \end{bmatrix}^{\top}$ as a new training set $\mathbf{X}_{\textup{N}}\in\mathbb{R}^{(p+c) \times n}$ and $\begin{bmatrix} \mathbf{D}^{\top} , \sqrt{\gamma}\;\mathbf{W}^{\top}\end{bmatrix}^{\top}$ as a new dictionary $\mathbf{D}_{\textup{N}}\in\mathbb{R}^{(p+c) \times k}$, (\ref{eq:DK-SVD}) is converted to the \emph{lasso}
\begin{equation}\label{eq:modifiedLasso}
\min_{\mathbf{D}_{\textup{N}},\mathbf{A}}\frac{1}{2}\|\mathbf{X}_{\textup{N}}-\mathbf{D}_{\textup{N}}\mathbf{A}\|_{\textup{F}}^{2}+\lambda\|\mathbf{A}\|_{1},
\end{equation}
and can efficiently be solved by one of the recently developed fast algorithms for this purpose such as K-SVD~\cite{DL:Aharon06}. Deriving $\mathbf{D}$ and $\mathbf{W}$ from $\mathbf{D}_{\textup{N}}$ is straightforward and the details are provided in~\cite{DL:QiangZhang10}.

\subsubsection{Label consistent K-SVD (LCK-SVD)}

Inspired by the DK-SVD as described in previous subsection, Jiang \emph{et al.}~\cite{DL:Jiang11} proposed label consistent K-SVD (LCK-SVD). In DK-SVD, although the linear classifier $\mathbf{W}$ and dictionary $\mathbf{D}$ are learned in one optimization problem, there is no mechanism to ensure that the dictionary learned is discriminative. To overcome this problem, it is suggested in~\cite{DL:Jiang11} to enforce a label consistency constraint on the dictionary by adding one additional term to the optimization problem of DK-SVD given in~(\ref{eq:DK-SVDPre}). The LCK-SVD optimization problem is, therefore, as follows:
\begin{equation}\label{eq:LC-KSVDPre}
\min_{\substack{\mathbf{D},\mathbf{A} \\ \mathbf{W},\mathbf{A}}}\frac{1}{2}\|\mathbf{X}-\mathbf{D}\mathbf{A}\|_{\textup{F}}^{2}+\frac{\eta}{2}\|\mathbf{Q}-\mathbf{U}\mathbf{A}\|_{\textup{F}}^{2}+ \frac{\gamma}{2}\|\mathbf{Y}-\mathbf{W}\mathbf{A}\|_{\textup{F}}^{2}+\lambda\|\mathbf{A}\|_{1},
\end{equation}
where the added second term enforces the label consistency on the dictionary. In other words, the second term in~(\ref{eq:LC-KSVDPre}) enforces the coefficients $\mathbf{A}$ to be as similar as possible to the optimal discriminative sparse codes in $\mathbf{Q}$. In~(\ref{eq:LC-KSVDPre}), $\mathbf{Q}\in \mathbb{R}^{k\times n}$ is encoding the optimal discriminative sparse coefficients, $\mathbf{U}\in \mathbb{R}^{k\times k}$ is a linear transformation matrix, and $\eta$ is a parameter that controls the relative contribution of the label consistency term. Each column of $\mathbf{Q}$ is $\mathbf{q}_{i}=\{0,...,1,1,...,0\}^{\top}$, where the locations of ones correspond to the optimal nonzero sparse coefficients representing a data sample $\mathbf{x}_i$. For example, if both $\mathbf{X}$ and $\mathbf{D}$ consist of six columns (six data samples and six dictionary atoms), such that there are two vectors in a three-class problem, $\mathbf{Q}$ has to be defined as:
\begin{equation}\label{eq:QExample}
\mathbf{Q}=\begin{bmatrix}
1 & 1 & 0 & 0 & 0 & 0\\
1 & 1 & 0 & 0 & 0 & 0\\
0 & 0 & 1 & 1 & 0 & 0\\
0 & 0 & 1 & 1 & 0 & 0\\
0 & 0 & 0 & 0 & 1 & 1\\
0 & 0 & 0 & 0 & 1 & 1
\end{bmatrix}.
\end{equation}

Similar to DK-SVD, the first three terms in~(\ref{eq:LC-KSVDPre}) can be combined into one term as follows:
\begin{equation}\label{eq:LC-KSVD}
\min_{\mathbf{D},\mathbf{A},\mathbf{W},\mathbf{U}}\frac{1}{2}\begin{Vmatrix}\begin{bmatrix}
\mathbf{X} \\ \sqrt{\eta}\;\mathbf{Q}   \\ \sqrt{\gamma}\;\mathbf{Y}
\end{bmatrix}-\begin{bmatrix}
\mathbf{D} \\ \sqrt{\eta}\;\mathbf{U} \\ \sqrt{\gamma}\;\mathbf{W}
\end{bmatrix}\mathbf{A}\end{Vmatrix}_{\textup{F}}^{2} + \lambda\|\mathbf{A}\|_{1}.
\end{equation}
Let $\begin{bmatrix} \mathbf{X}^{\top} , \sqrt{\eta}\;\mathbf{Q}^{\top}, \sqrt{\gamma}\;\mathbf{Y}^{\top} \end{bmatrix}^{\top}$ be a new training set $\mathbf{X}_{\textup{N}}\in\mathbb{R}^{(p+k+c) \times n}$ and $\begin{bmatrix} \mathbf{D}^{\top}, \sqrt{\eta}\;\mathbf{U}^{\top}, \sqrt{\gamma}\;\mathbf{W}^{\top}\end{bmatrix}^{\top}$ be a new dictionary $\mathbf{D}_{\textup{N}}\in\mathbb{R}^{(p+k+c) \times k}$, (\ref{eq:LC-KSVD}) is converted to the form given in~(\ref{eq:modifiedLasso}), which can be again efficiently solved by one of the recently developed fast algorithms for this purpose such as K-SVD~\cite{DL:Aharon06}. Subsequently, $\mathbf{D}$, $\mathbf{U}$, and $\mathbf{W}$ can be easily derived from $\mathbf{D}_{\textup{N}}$.

\subsubsection{Bayesian supervised dictionary learning}

Dictionary learning based on Bayesian models was first proposed by Zhou \emph{et al.}~\cite{DL:Zhou09,DL:Zhou12}. However, the method did not take into account the class labels in learning the dictionary and hence, was not optimal for classification tasks. In order to overcome this problem, recently, a non-parametric Bayesian technique has been proposed to jointly learn the dictionary, classifier, and sparse coefficients using beta-Bernoulli process~\cite{DL:Babagholami13}.

\paragraph*{Discussion} The idea used in this category of S-DLSR is more sophisticated than the previous two. However, the major disadvantage especially with the first approach in this category, i.e., SDL-D, is that the optimization problem is non-convex and complex. If the optimization is performed alternately between learning the dictionary and classifier parameters, it is quite likely to become stuck in some local minima. On the other hand, due to the complexity of the optimization problem (except for the bilinear classifier in~\cite{DL:Mairal08a}), linear classifiers are merely considered in this category, which are usually too simple to solve difficult classification tasks, and can only be successful in simple ones as shown in~\cite{DL:Mairal08a}. Another major problem with the approaches in this category of S-DLSR is that there exist many parameters involved in the formulation, which are hard and time-consuming to tune (see for example~\cite{DL:Mairal08a,DL:Mairal12}).

\subsection{Embedding Class Labels into the Learning of Dictionary}\label{subsec:CatInfoDL}

The fourth category of S-DLSR approaches includes the category information in the learning of the dictionary. Among the approaches in this category, Gangeh \emph{et al.}~\cite{Gangeh:TSP13} and Zhang \emph{et al.}~\cite{DL:Zhang13} have proposed to learn the dictionary and sparse coefficients in a more discriminative (in some sense) projected space, whereas Lazebnik and Raginsky~\cite{DL:Lazebnik09} included the category information into the learning of the dictionary by minimizing the information loss due to predicting the class labels in the space of the learned dictionary instead of the original space. The details of these methods are as follows.

\subsubsection{HSIC-based supervised dictionary learning}

Recently, Gangeh \emph{et al.}~\cite{Gangeh:TSP13} proposed an S-DLSR method based on Hilbert Schmidt independence criterion (HSIC). HSIC is a kernel-based independence measure between two random variables $\mathcal{X}$ and $\mathcal{Y}$~\cite{HSIC:Gretton05b}. It computes the Hilbert-Schmidt norm of the cross-covariance operators in reproducing kernel Hilbert Spaces (RKHSs)~\cite{HSIC:Gretton05b,HSIC:Aronszajn50}.

In practice, HSIC is estimated using a finite number of data samples. Let $\mathcal{Z}:=\{(\mathbf{x}_1,\mathbf{y}_1,),...,(\mathbf{x}_n,\mathbf{y}_n)\}\subseteq\mathcal{X}\times\mathcal{Y}$ be $n$ independent observations drawn from $p:=P_{\mathcal{X}\times\mathcal{Y}}$. The empirical estimate of HSIC can be computed using~\cite{HSIC:Gretton05b}
\begin{equation}\label{eq:HSICEmp}
\textup{HSIC}(\mathcal{Z})=\frac{1}{(n-1)^2}\textup{tr}(\mathbf{KHLH}),
\end{equation}
where $\textup{tr}$ is the trace operator, $\mathbf{H}, \mathbf{K}, \mathbf{L}\in\mathbb{R}^{n\times n}, K_{i,j}=k(x_i,x_j), L_{i,j}=l(y_i,y_j)$, and $\mathbf{H}=\mathbf{I}-n^{-1}\mathbf{ee}^\top$ ($\mathbf{I}$ is the identity matrix,  and $\mathbf{e}$ is a vector of $n$ ones, and hence, $\mathbf{H}$ is the centering matrix). According to~(\ref{eq:HSICEmp}), maximizing the empirical estimate of HSIC, i.e., $\textup{tr}(\mathbf{KHLH})$, will lead to the maximization of the dependency between two random variables $\mathcal{X}$ and $\mathcal{Y}$.

The HSIC-based S-DLSR learns the dictionary in a space where the dependency between the data and corresponding class labels is maximized. To this end, it has been proposed in~\cite{Gangeh:TSP13} to solve the following optimization problem
\begin{equation}\label{eq:SDLHSICConst}
\begin{aligned}
& \underset{\mathbf{U}}{\text{max}}
& & \text{tr}(\mathbf{U}^{\top}\mathbf{XHLH}\mathbf{X}^{\top}\mathbf{U}), \\
& \text{s.t.}
& & \mathbf{U}^{\top}\mathbf{U}=\mathbf{I}
\end{aligned}
\end{equation}
where $\mathbf{X} = [\mathbf{x}_1,\mathbf{x}_2,...,\mathbf{x}_n]\in\mathbb{R}^{p \times n}$ is $n$ data samples with the dimensionality of $p$; $\mathbf{H}$ is the centering matrix, and its function is to center the data, i.e., to remove the mean from the features; $\mathbf{L}$ is a kernel on the labels $\mathbf{Y}$; and $\mathbf{U}$ is the transformation that maps the data to the space of maximum dependency with the labels. According to the Rayleigh-Ritz Theorem~\cite{book:Lutkepohl96}, the solution for the~(\ref{eq:SDLHSICConst}) is the top eigenvectors of $\bm{\Phi}=\mathbf{XHLHX}^{\top}$ corresponding to its largest eigenvalues.

To explain how the optimization problem provided in~(\ref{eq:SDLHSICConst}) learns the dictionary in the space of maximum dependency with the labels, using a few manipulations, we note that the objective function given in~(\ref{eq:SDLHSICConst}) has the form of empirical HSIC given in~(\ref{eq:HSICEmp}), i.e.,
\begin{align}\label{eq:SDLToHSIC}    
  \makebox[2em][l]{$\underset{\mathbf{U}}{\text{max}}\; \text{tr}(\mathbf{U}^{\top}\mathbf{XHLH}\mathbf{X}^{\top}\mathbf{U})$}  \nonumber \\
  &=\underset{\mathbf{U}}{\text{max}}\; \text{tr}(\mathbf{X}^{\top}\mathbf{U}\mathbf{U}^{\top}\mathbf{XHLH}) \nonumber \\
  &=\underset{\mathbf{U}}{\text{max}}\; \text{tr}\bigg(\bigg[(\mathbf{U}^{\top}\mathbf{X})^{\top}\mathbf{U}^{\top}\mathbf{X}\bigg]\mathbf{HLH}\bigg) \nonumber \\
  &=\underset{\mathbf{U}}{\text{max}}\; \text{tr}(\mathbf{KHLH}),
\end{align}
where $\mathbf{K}=(\mathbf{U}^{\top}\mathbf{X})^{\top}\mathbf{U}^{\top}\mathbf{X}$ is a linear kernel on the transformed data in the subspace $\mathbf{U}^{\top}\mathbf{X}$. To derive~(\ref{eq:SDLToHSIC}), it is noted that the trace operator is invariant under cyclic permutation.

Now, it is easy to observe that the form given in~(\ref{eq:SDLToHSIC}) is the same as the empirical HSIC in~(\ref{eq:HSICEmp}) up to a constant factor and therefore, it can be easily interpreted as transforming centered data $\mathbf{X}$ using the transformation $\mathbf{U}$ to a space where the dependency between the data and class labels is maximized. In other words, the computed transformation $\mathbf{U}$ constructs the dictionary learned in the space of maximum dependency between the data and class labels.

After finding the dictionary $\mathbf{D}=\mathbf{U}$, the sparse coefficients can be computed using the formulation given in~(\ref{eq:lasso})~\cite{Gangeh:TSP13}.

One main advantage of the HSIC-based S-DLSR is that both dictionary and sparse coefficients can be computed in closed form~\cite{Gangeh:TSP13}, which makes the approach computationally very efficient. Another main advantage of the approach is that it can be easily kernelized and therefore, by embedding an appropriate kernel into the solution, subtle classification tasks can be solved with high accuracy. The approach, however, does not allow overcomplete dictionaries due to the orthogonality constraint imposed on the transformation. This might be of little concern as it has been shown that the method works comparably well at small dictionary sizes~\cite{Gangeh:TSP13}.

\subsubsection{Discriminative projection and dictionary learning}

In the same line as HSIC-based S-DLSR, Zhang \emph{et al.}~\cite{DL:Zhang13} also proposed to learn the dictionary and the sparse representation in a more discriminative (in some sense, which will be defined in next lines) space. To this end, they propose to first project the data to an orthogonal space where the intra- and inter-class reconstruction errors are minimized and maximized, respectively, and subsequently learn the dictionary and the sparse representation of the data in this space. Intra-class reconstruction error for a data sample $\mathbf{x}_i$ is defined as the reconstruction error using the dictionary atoms in the ground-truth class of $\mathbf{x}_i$ under the metric $\mathbf{UU}^{\top}$ ($\mathbf{U}$ is the projection to be learned), whereas inter-class error is defined as the reconstruction error using the dictionary atoms other than the ground-truth class of $\mathbf{x}_i$ under the same metric.

To provide the mathematical formulation, given a set of training set $\mathbf{X}\in \mathbb{R}^{p\times n}$, the task is to learn a discriminative trasnformation/projection $\mathbf{U}\in \mathbb{R}^{p\times m}$, where $m\leq p$ is the number of basis, and dictionary $\mathbf{D}\in \mathbb{R}^{p\times k}$ using the optimization problem given below
\begin{equation}\label{eq:DSRC}
\begin{aligned}
& \underset{\mathbf{U},\mathbf{D}}{\text{min}}
& & \frac{1}{n}\sum_{i=1}^{n}\begin{pmatrix} S_{\beta}(R(\mathbf{x}_i))+\lambda\| \bm{\alpha}_i \|_1 \end{pmatrix} \\
& \text{s.t.}
& & \mathbf{U}^{\top}\mathbf{U}=\mathbf{I}
\end{aligned}
\end{equation}
where $S_{\beta}(x)=\frac{1}{1+e^{\beta(1-x)}}$ is a sigmoid function centered at 1 with the slope of $\beta$, and $R(\mathbf{x}_i)$ is the ratio of intra- to inter-class reconstruction errors. $S_{\beta}(R(\mathbf{x}_i))$ can be intuitively considered as the inverse classification confidence and by minimizing this term over the training samples in the objective function of~(\ref{eq:DSRC}), the discriminative projections $\mathbf{U}$ and dictionary $\mathbf{D}$ are empirically learned subject to a sparsity constraint imposed as the second term in~(\ref{eq:DSRC}).

In~(\ref{eq:DSRC}), $\bm{\alpha}_i$ is the sparse representation of the projected data sample $\mathbf{U}^{\top}\mathbf{x}_i$ in the space of dictionary learned in the projected space $\mathbf{U}^{\top}\mathbf{D}$, i.e.,
\begin{equation}\label{eq:CoeffDSRC}
\hat{\bm{\alpha}}_i=\min_{\bm{\alpha}_i}\begin{pmatrix}\|\mathbf{U}^{\top}\mathbf{x}_i-\mathbf{U}^{\top}\mathbf{D}\bm{\alpha}_i\|_{2}^{2}+\lambda\|\bm{\alpha}_i\|_{1}\end{pmatrix}.
\end{equation}

The optimization problem given in~(\ref{eq:DSRC}) and~(\ref{eq:CoeffDSRC}) has to be solved alternately between sparse coding (using ~(\ref{eq:CoeffDSRC}) with $\mathbf{U}$ and $\mathbf{D}$ fixed) and learning the dictionary and projected space (using~(\ref{eq:DSRC}) with fixed sparse coefficients $\mathbf{A}$). This optimization problem is non-convex and the projection and dictionary have to be learned iteratively and alternately using gradient descent. Therefore, unlike HSIC-based S-DLSR, there exist no closed-form solutions here and the algorithm may get stock in some local minima.

\subsubsection{Information loss minimization (info-loss)}

Lazebnik and Raginsky proposed in~\cite{DL:Lazebnik09} to include category information into the learning of the dictionary, by minimizing the information loss due to predicting labels from a supervised dictionary learned instead of original training data samples. This approach is known as \emph{info-loss} in the S-DLSR literature. In fact, in S-DLSR, the ultimate goal is to represent the original high-dimensional feature space by a dictionary such that it can facilitate the prediction of the class labels correctly. Ideally, the dictionary should maintain all discriminative power of the original feature space. However, some of this information is lost during the quantization of the feature space. In~\cite{DL:Lazebnik09}, it has been proposed to learn the dictionary such that the information loss
\begin{equation}\label{eq:infoloss}
I(\mathcal{X},\mathcal{Y})-I(\mathcal{D},\mathcal{Y})
\end{equation}
is minimized, where $I$ indicates the mutual information between its arguments as random variables, and $\mathcal{X}$, $\mathcal{D}$, and $\mathcal{Y}$ are the random variables on the original feature space $\mathbf{X}$, learned dictionary $\mathbf{D}$, and the class labels $\mathbf{Y}$, respectively.

Just the same as in the previous category of S-DLSR, the info-loss approach has the major drawback that it may become stuck in local minima. This is mainly because the optimization has to be done iteratively and alternately on two updates, as there is no closed-form solution for the approach. 

\subsubsection{Randomized clustering forests (RCF\nomenclature{RCF}{Randomized Clustering Forest})}

In~\cite{DL:Moosmann06}, it is proposed to learn the dictionary atoms using extremely randomized decision trees. This approach also falls into the second category of SDLs, as it seems that it starts from a very large dictionary using random forests, and tries to prune it later to conclude with a smaller dictionary.

\paragraph*{Discussion} The idea of learning the dictionary and sparse coefficients in a more discriminative projected space introduced by the first two approaches in the category, i.e., HSIC-based S-DLSR and discriminative projection and dictionary learning opens a very promising avenue of research in the field of S-DLSR. Based on this two methods, the projection to a discriminative space can be defined in different ways depending on some criteria related to the problem at hand. If the projection/dictionary are defined to be orthonormal, the learning of the coefficients can be performed in closed form~\cite{Gangeh:TSP13,DL:Friedman07} using soft-thresholding~\cite{DL:Donoho95}. With a careful selection of the discriminative criterion, it might be also possible to find a closed-form solution for the dictionary such as the one found in HSIC-based S-DLSR that can further improve the performance of the approach in terms of computation time.

\subsection{Embedding Class Labels into the Learning of Sparse Coefficients}\label{subsec:CatInfoCoeff}

The fifth category of S-DLSR includes class category in the learning of coefficients~\cite{DL:Huang07} or in the learning of both dictionary and coefficients~\cite{DL:Rodriguez07,DL:MengYang11}. Supervised coefficient learning in all these papers~\cite{DL:Huang07,DL:Rodriguez07,DL:MengYang11} has been performed more or less in the same way using the Fisher discrimination criterion~\cite{DIM:Fisher36}, i.e., by minimizing the within-class covariance of coefficients and at the same time maximizing their between-class covariance. As for the dictionary, while Huang \emph{et al.}~\cite{DL:Huang07} have used predefined basis by deploying an overcomplete dictionary as a combination of Haar and Gabor basis, Yang \emph{et al.}~\cite{DL:MengYang11} have proposed a discriminative fidelity term to learn the dictionary, for which further description is provided below, along with the learning of the coefficients.

\subsubsection{Fisher discrimination dictionary learning (FDDL\nomenclature{FDDL}{Fisher Discrimination Dictionary Learning})}

In~\cite{DL:MengYang11}, an approach called Fisher discrimination dictionary learning (FDDL) has been proposed, that uses category information in learning both dictionary and sparse coefficients. To learn the dictionary supervised, a discriminative fidelity term has been proposed that encourages learning dictionary atoms of one class from the training samples of the same class, and at the same time penalizes their learning by the training samples from other classes. As stated above, the coefficients have been learned supervised, by including the Fisher discriminant criterion in their learning.

To provide a mathematical formulation for FDDL, suppose that the training samples are grouped according to the classes they belong to, i.e., $\mathbf{X} = [\mathbf{X}_1,\mathbf{X}_2,...,\mathbf{X}_c]\in\mathbb{R}^{p \times n}$, where $c$ is the number of classes. The objective function in FDDL consists of two terms: a fidelity term and a discrimination constraint term on coefficients
\begin{equation}\label{eq:FDDLTop}
J(\mathbf{D},\mathbf{A})=\underset{\mathbf{D},\mathbf{A}}{\text{min}}\;\;\;
r(\mathbf{X},\mathbf{D},\mathbf{A}) + \lambda_1\left \| \mathbf{A} \right \|_1 + \lambda_2f(\mathbf{A}),
\end{equation}
where $r(\mathbf{X},\mathbf{D},\mathbf{A})$ is the fidelity term and $f(\mathbf{A})$ is the discrimination constraint on the coefficients.

The fidelity term is defined in~\cite{DL:MengYang11} as follows
\begin{equation}\label{eq:FDDLFidelity}
r(\mathbf{X},\mathbf{D},\mathbf{A}) = \left \| \mathbf{X}_i-\mathbf{D}\mathbf{A}_i \right \|_{\textup{F}}^2 + \left \| \mathbf{X}_i-\mathbf{D}_i\mathbf{A}_i^i \right \|_{\textup{F}}^2 + \sum_{\substack{j=1\\j\neq i}}^{c}\left \| \mathbf{D}_j \mathbf{A}_i^j \right \|_{\textup{F}}^2,
\end{equation}
where $\mathbf{D}_i$ is the part of the dictionary associated with class $i$, and $\mathbf{A}_i$ is the representation of $\mathbf{X}_i$ over $\mathbf{D}$. Also $\mathbf{A}_i = [\mathbf{A}_i^1,\mathbf{A}_i^2,...,\mathbf{A}_i^c]$, where $\mathbf{A}_i^j$ is the part of the coefficients that represent $\mathbf{X}_i$ over the subdictionary $\mathbf{D}_j$. In~(\ref{eq:FDDLFidelity}), the first two terms indicate that the whole dictionary and also the subdictionary associated with class $i$ should well represent the data samples in the same class $\mathbf{X}_i$, whereas the last term indicates that the subdictionaries from other classes have little contribution towards the representation of the data samples in class $i$.

The Fisher discrimination term, on the other hand, is as follows
\begin{equation}\label{eq:FDDLFisher}
f(\mathbf{A})=\textup{tr}(S_{\textup{W}}(\mathbf{A})) - \textup{tr}(S_{\textup{B}}(\mathbf{A})) + \eta\left \|\mathbf{A}  \right \|_{\textup{F}}^2,
\end{equation}
where $\textup{tr}$ is the trace operator; $S_{\textup{W}}$ and $S_{\textup{B}}$ are within- and between-class covariance matrices, respectively. The last term is a penalty added to~(\ref{eq:FDDLFisher}) to make the optimization problem convex~\cite{DL:MengYang11}.

\paragraph*{Discussion} The joint optimization problem, due to the Fisher discrimination criterion on the coefficients and the discriminative fidelity term on the dictionary proposed in~(\ref{eq:FDDLTop}), is not convex, and has to be solved iteratively and alternately between these two terms until it converges. However, there is no guarantee to find the global minimum. Also, it is not clear whether the improvement obtained in classification by including the Fisher discriminant criterion on coefficients justifies the additional computation load imposed on the learning, as there is no comparison provided in~\cite{DL:MengYang11} on the classification with and without including supervision on coefficients.

\subsection{Learning a Histogram of Dictionary Elements over Signal Constituents}\label{subsec:histogram}

There are situations where a signal is made of some local constituents, e.g., an image is made up of patches or a speech, which is consisting of phonemes. However, the ultimate classification task is to classify the signal, not its individual local constituents, e.g., the whole image, not the patches in the previous example. This classification task is usually tackled by computing the histogram of dictionary atoms computed over local constituents of a signal. The computed histograms are used as the signature (model) of the signal, which are eventually used for the training of a classifier and predicting the labels of unknown signals. Unlike the previous five categories, the motivation of the approaches in the sixth S-DLSR category is to design a supervised dictionary, which is discriminative over the histogram representation of signals, not over individual local descriptors~\cite{DL:Lian10,DL:WeiZhang09,DL:Perronnin08}. Hence, these approaches cannot be used in cases where a signal does not consist of a collection of local constituents. The main approaches in this category are: 1) texton-based method, 2) histogram computation using DLSR, 3) universal and adapted vocabularies, and 4) supervised dictionary learning model (SDLM). The following subsections provide the description of these methods.

\subsubsection{Texton-based approach}\label{subsubsec:texton}

The texton-based approach~\cite{texton:Julesz81,texton:Leung01,texton:Cula04,texton:Schmid04,texton:varma05,texton:varma09}, is one of the earliest methods that was proposed to compute the histogram of dictionary elements, called textons, to model a texture image based on patches extracted. This approach was particularly proposed for texture analysis, but is sufficiently general to be used in other applications. In a texton-based approach, the first step is to construct the dictionary. To this end, small-sized local patches are randomly extracted from each texture image in the training set. These small patches are then aggregated over all images in a class, and clustered using a clustering algorithm such as \emph{k}-means. Obtained cluster centers form a dictionary that represents the class of textures used. In other words, supervised \emph{k}-means is used to compute the dictionary atoms~\cite{texton:varma05,texton:varma09}.

\begin{figure*}[!tb]
  \centering
  \subfloat[ ]{\label{fig:textonDict} \includegraphics[width=.495\textwidth]{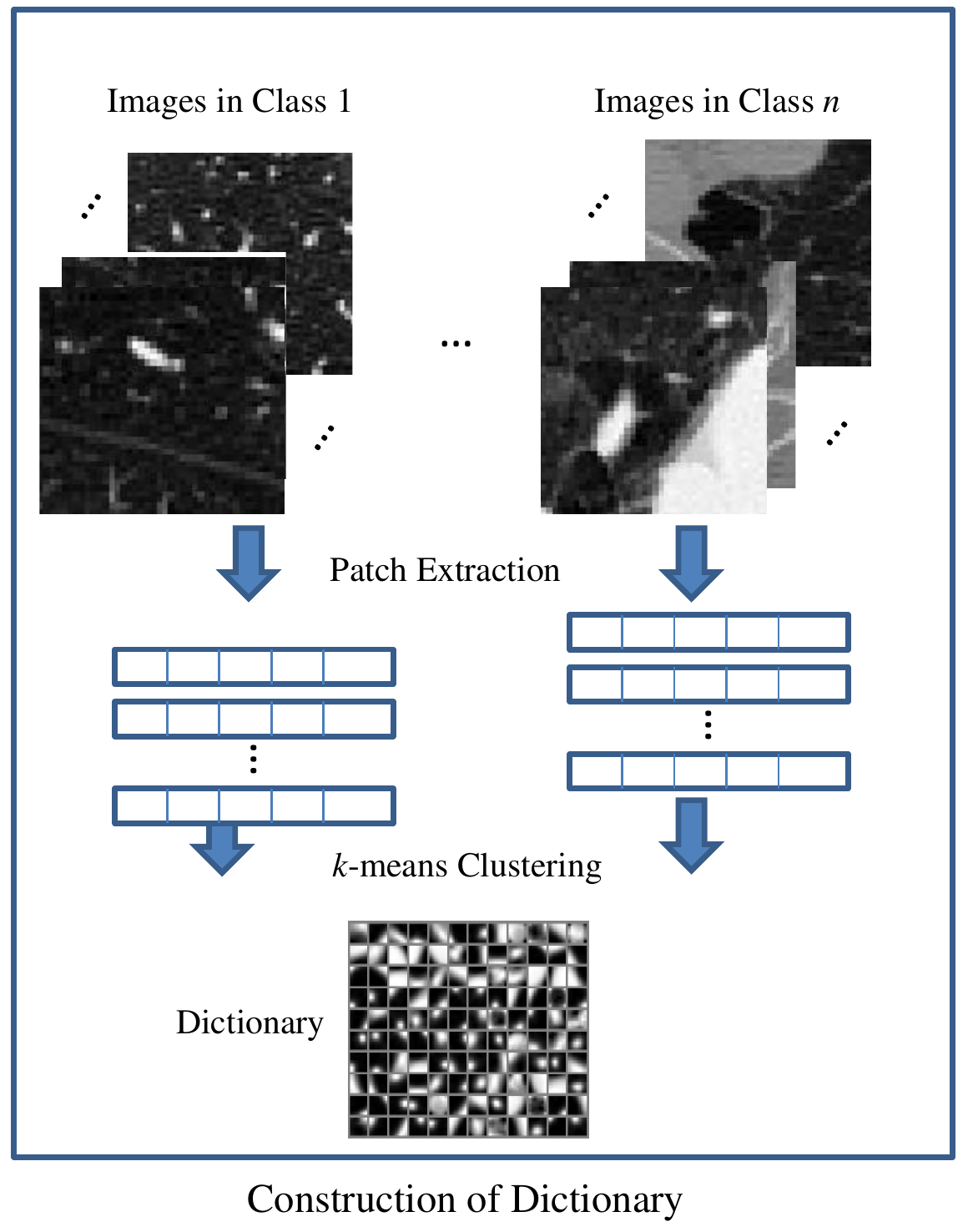}}
  \subfloat[ ]{\label{fig:textonHist} \includegraphics[width=.45\textwidth]{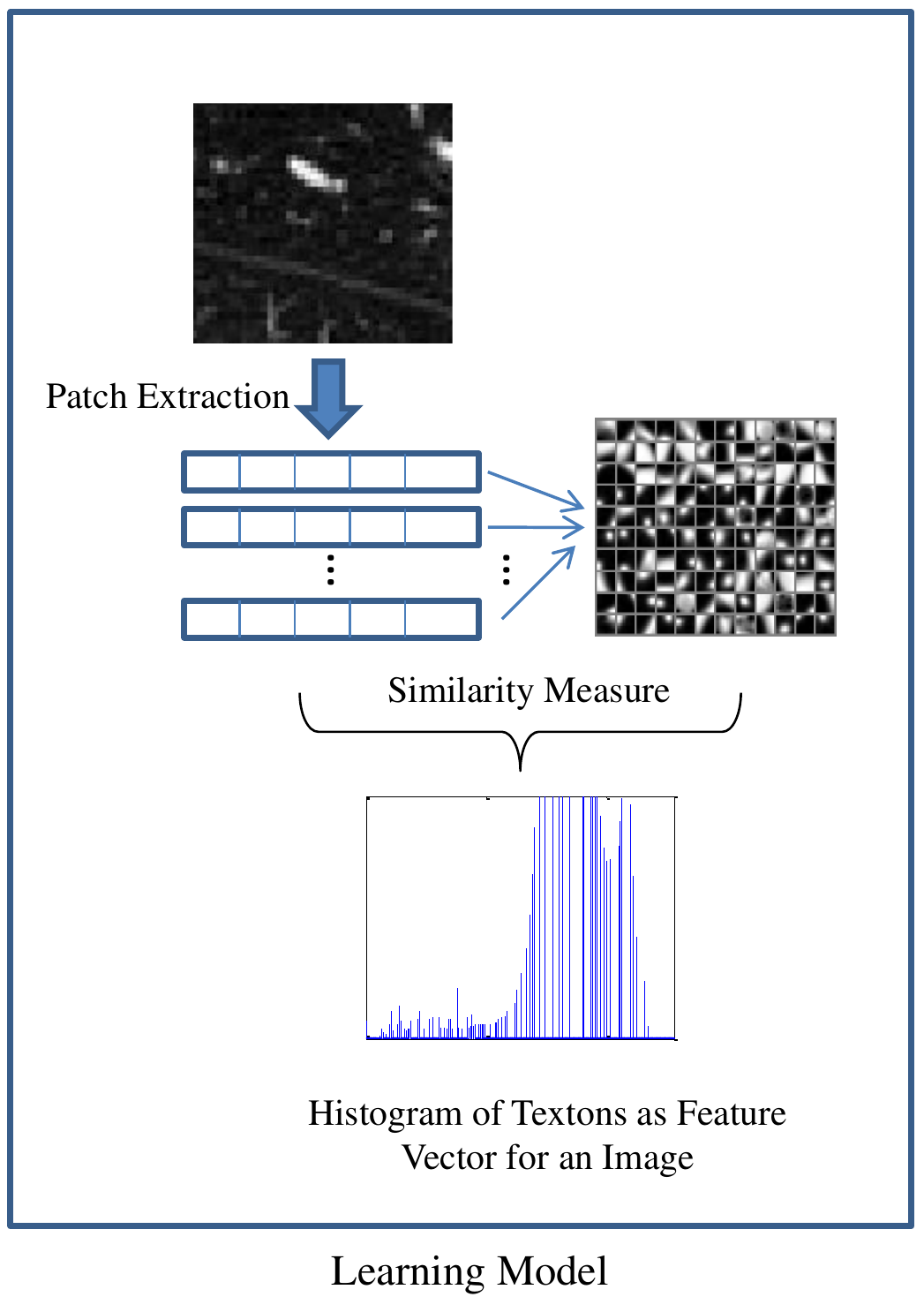}}
  \caption[The illustration of two steps of a texton-based system.]{The illustration of two steps of a texton-based system: (a) the generation of texton dictionary using supervised \emph{k}-means (b) and the generation of features by computing the texton histograms on an image (reused from~\cite{Gangeh:MICCAI10} courtesy of Springer Science).}
  \label{fig:textonModel}
\end{figure*}

The next step is to find the features (learn the model) using the images in the training set. To this end, small patches of the same size as the previous step are extracted by sliding a window over each training image in a class. Then the distance between each patch to all textons in the dictionary are computed, to find the closest match using a distance measure such as Euclidean distance. Finally, a histogram of textons is updated accordingly for each image based on the closest match found. This yields a histogram for each image in the training set, which is used as the features representing that image after normalization. Figure~\ref{fig:textonModel} illustrates the construction of the dictionary and learning of the model in a texton-based system.

\subsubsection{Histogram computation using dictionary learning and sparse representation}

In the texton-based approach, supervised \emph{k}-means was used to compute the dictionary. To compute the histogram of textons, each patch was represented by the closest match in the dictionary. This is the maximum sparsity possible as each patch is represented by only one dictionary element. However, as proposed in~\cite{Gangeh:ICIAR11}, it is possible to use~(\ref{eq:lasso}) and one of the recent algorithms for its implementation, such as online learning~\cite{DL:Mairal10}, to compute the dictionary and the corresponding sparse coefficients over the patches extracted from an image. The same as the texton-based approach, building the dictionary and histogram of dictionary elements can be done in two steps.

In the first step, random patches are extracted from each image in the training set. Next, by submitting these patches into the online learning algorithm, the dictionary can be computed~\cite{Gangeh:ICIAR11}.

As the second step, it is needed to find the model (feature set) for each image. To this end, patches of the same size as those in the dictionary learning step are extracted from each image. Let $\mathbf{x}_i$ be the $i^{\textup{th}}$ image in the training set. The signal constituents (i.e., patches) of $\mathbf{x}_i$ can be denoted as $\mathbf{x}_i=[\mathbf{x}_i^1,\mathbf{x}_i^2,...,\mathbf{x}_i^m]\in\mathbb{R}^{t\times m}$, where $m$ is the number of patches extracted, and each patch size is $\sqrt{t}\times \sqrt{t}$. Then using~(\ref{eq:lasso}), the corresponding coefficients $\bm{\alpha}_i=[\bm{\alpha}_i^1,\bm{\alpha}_i^2,...,\bm{\alpha}_i^m]\in\mathbb{R}^{k \times m}$ are computed ($k$ is the number of dictionary atoms). For each patch $\mathbf{x}_i^j$, most of the elements in the corresponding coefficient $\bm{\alpha}_i^j$ are zero. The nonzero elements in $\bm{\alpha}_i^j$ determine the atoms in the dictionary $\mathbf{D}$ that contribute towards the representation of the patch $\mathbf{x}_i^j$. If all these coefficients are summed up for all patches extracted from an image, one can effectively find the histogram of primitive elements contributing towards the representation of this particular image, i.e.,
\begin{equation}\label{eq:coefHist}
\mathbf{h}(\mathbf{x}_i)=\sum_{j=1}^m \bm{\alpha}_i^j.
\end{equation}
A histogram $\mathbf{h}$ with positive values in all bins can be eventually obtained by imposing a positive constraint on $\bm{\alpha}_i^j$ in~(\ref{eq:lasso}). The positive constraint also prevents canceling the effect of different patches when they are summed up in~(\ref{eq:coefHist}). 

In this way, while in a texton-based approach each patch is represented using only the closest texton in the dictionary, here each patch is represented by using several primitive elements in the dictionary, and hence can potentially provide richer representation than the texton-based approach. The number of nonzero elements in $\bm{\alpha}_i^j$, and consequently in $\bm{\alpha}_i$, can be controlled using $\lambda$, which is the sparsity parameter in~(\ref{eq:lasso}), i.e., larger values of $\lambda$ yield sparser coefficients~\cite{DL:Mairal10}.

\begin{figure*}[!tb]
  \centering
  \includegraphics[width=.95\textwidth]{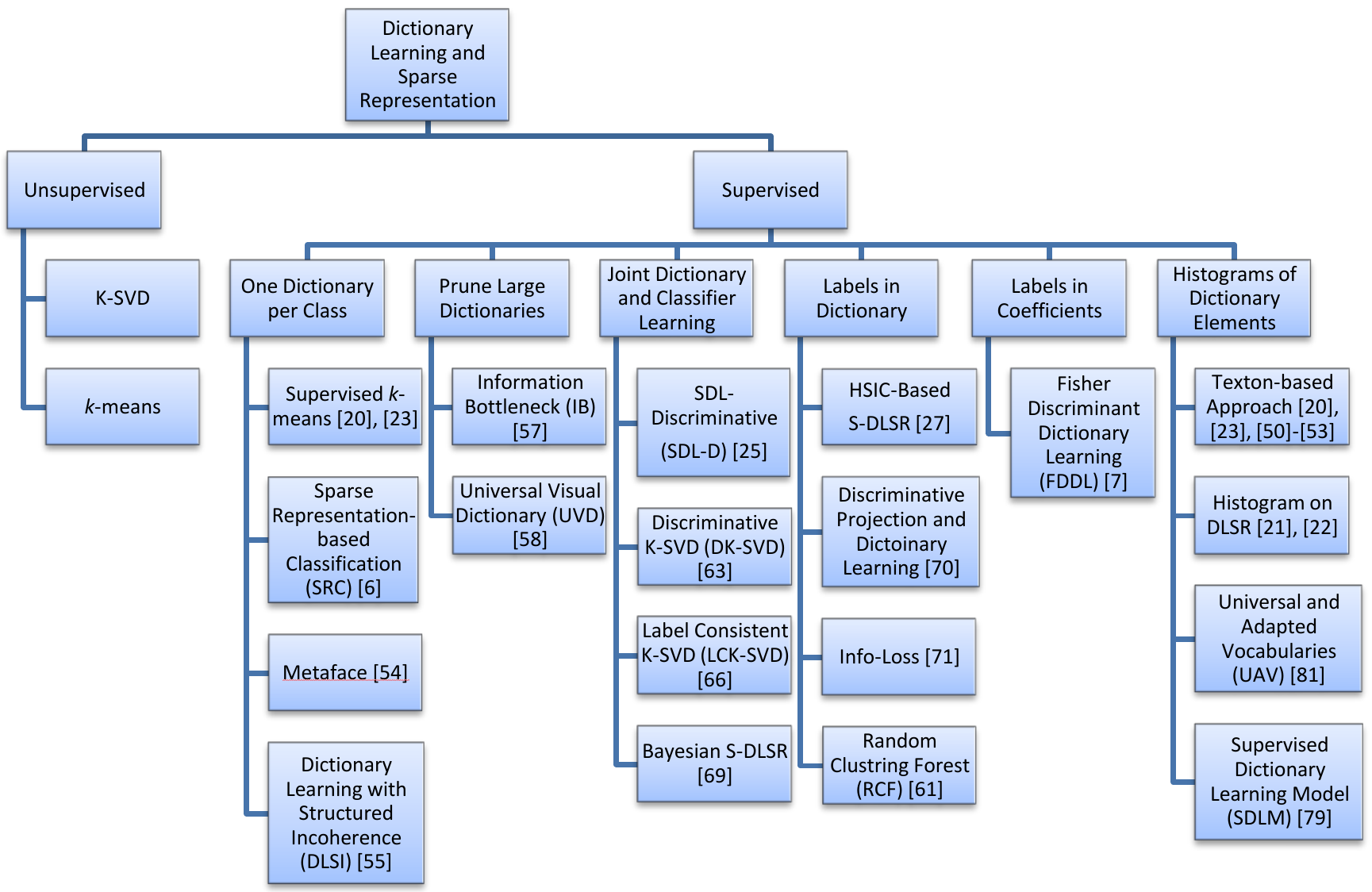}
  \caption[Taxonomy of dictionary learning and sparse representation approaches.]{Taxonomy of dictionary learning and sparse representation as presented in this paper. Supervised dictionary learning and sparse representation (S-DLSR) approaches are divided into six categories.}
  \label{fig:Taxonomy}
\end{figure*}

\subsubsection{Universal and adapted vocabularies (UAV\nomenclature{UAV}{Universal and Adapted Vocabulary})}

Although the above two approaches include the category information into the learning of individual dictionary atoms, they do not include the class labels into the learning of the histograms. This is while the main goal is to make the histogram discriminative not the individual dictionary elements as the ultimate goal is to classify the signal not its constituents. For example, a white patch may appear in outdoor scenes as part of a cloud in sky as well as on indoor scenes as the color on the ceiling of a kitchen. However, the main goal is to classify the scenes to indoors and outdoors and hence putting some efforts to make the individual dictionary elements discriminative might be misleading (a white dictionary atom may appear in both outdoor and indoor scenes in previous example). To address these kind of problems, Perronnin has proposed in~\cite{DL:Perronnin08} to learn one bipartite histogram per class for each image. Each bipartite histogram, as the name implies, has two parts: a part adapted to the specific class, and a universal part. In each histogram, ideally, if the object belongs to the class, its adapted part is more significant than the universal one; otherwise the universal part is more dominant.

Gaussian mixture models (GMM\nomenclature{GMM}{Gaussian Mixture Model}) are used to learn the universal vocabularies (dictionaries) using maximum likelihood estimation (MLE\nomenclature{MLE}{Maximum Likelihood Estimation}) for low level local descriptors such as scale-invariant feature transform (SIFT\nomenclature{SIFT}{Scale Invariant Feature Transform}) descriptors. Then class specific vocabularies are adapted by the maximum a posteriori (MAP\nomenclature{MAP}{Maximum A Posteriori}) criterion. Eventually, the bipartite histograms are estimated by using the adapted and universal vocabularies~\cite{DL:Perronnin08}.

\subsubsection{Supervised dictionary learning model (SDLM\nomenclature{SDLM}{Supervised Dictionary Learning Model})}

A supervised dictionary learning model (SDLM) is proposed in~\cite{DL:Lian10}, which combines an unsupervised model based on a Gaussian mixture model (GMM) with a supervised model, i.e., a logistic regression model in a probabilistic framework. As explained in the beginning of this subsection, the motivation of this model is to learn the dictionary such that the histogram representation of images are sufficiently discriminative over different classes. Intuitively, in SDLM, a logistic loss function is used to pass the discriminative information in class labels to histogram features. This information is subsequently passed to the dictionary learned over image local features by affecting the GMM parameters~\cite{DL:Lian10}.

\paragraph*{Discussion} As mentioned earlier, the approaches in this category are mainly designed to classify signals using the histogram of dictionary atoms built on the signal constituents. Since the ultimate goal is to classify the signals per se not their constituents, it is reasonable to make the histograms discriminative not necessarily the individual dictionary atoms. The last two approaches in this category, i.e., UAV and SDLM  place more emphasis on this attribute and propose methods to include category information into the learning process such that the histogram of dictionary atoms represent the signals in the most discriminative way possible.

Figure~\ref{fig:Taxonomy} summarizes the taxonomy of dictionary learning and sparse representation techniques as presented in this paper for a quick reference.

\section{Summary and Guidelines for Practitioners}\label{sec:Summary}

This section summarizes the methods presented in Section~\ref{sec:litReviewSDL} from the perspective of how the different building blocks of an S-DLSR algorithm are represented and learned. This perspective allows for the readers who are interested in building an S-DLSR solution to evaluate different design decisions and select the best practices for the problem at hand.

There are three building blocks in an S-DLSR method: (1)
the dictionary $\mathbf{D}$, (2) the coefficients $\mathbf{A}$, and (3) the classifier parameters $\mathbf{W}$. The methods presented in Section~\ref{sec:litReviewSDL} vary in how they represent and learn these blocks. In the rest of this section, we summarize the different design decisions for the \emph{representation} and \emph{learning} of each of these blocks and comment on the advantages and shortcomings of each.

\subsection{The dictionary $\mathbf{D}$}
The dictionary $\mathbf{D}\in\mathbb{R}^{p\times k}$ consists of a set of $k$ atoms. Each atom $\mathbf{d}_{i}$ is a $p$-dimensional vector which can be \emph{represented} as:
\paragraph*{i) a single data instance} i.e., $\mathbf{d}_{i}=\mathbf{x}_{i}$. This representation was used by the seminal work of sparse representation-based classification (SRC) \cite{DL:Wright09}. If the size of the training samples is manageable, this representation is very efficient as no overhead is needed to form the dictionary atoms. In addition, in the case that the classification is based on the minimum residual error given in~(\ref{eq:SRCResidue1}), this representation is easily interpretable as the system user can inspect the dictionary elements which result in the minimum residual error, and understand how the classification decision was made. On the other hand, this approach is sensitive to noisy training instances as the dictionary atoms are the training samples. Moreover, if the number of training samples is large, this representation is inefficient as it will be computationally complex to decode new signals. It will also be infeasible to store and transfer a large dictionary of instances especially when the recognition system are to be deployed on a modest hardware such as that of portable devices. One possible way is to reduce the size of the dictionary by selecting a representative subset of the dictionary atoms. This can be done by representing atoms (i.e., columns of $\mathbf{D}$) as vectors in the space of features and select a subset of these vectors such that the reconstruction error of other atoms (or the matrix $\mathbf{D}$)  based on the selected vectors is minimized. This problem is formally known as column subset selection (CSS) and several research efforts have been conducted for solving this problem.  ~\cite{CS:Drineas06,CS:Farahat13,CS:Boutsidis09a,CS:Farahat14}.

\paragraph*{ii) a function of multiple data instances} 
This representation was used by the \emph{metaface} method~\cite{DL:MengYang10} as well as its extensions~\cite{DL:Ramirez:10}. The basic idea here is to learn a small dictionary in which each atom is a linear/nonlinear combination of many data instances. The main advantage of this representation is the simplicity of the approaches used to learn the dictionary. 
However, since the size of the dictionary linearly increases with the number of classes, these approaches may lead to large dictionaries when there exist many classes. In addition, if the dictionary atoms are dense combination of many data instances, it will be difficult to interpret their meaning and reason about the different classification decisions.

\paragraph*{iii) a function of signal constituents} 
An example of this representation is the texton-based approach (Subsection~\ref{subsubsec:texton}), where each atom is an average-like function of some constituents of the original signal. This representation is useful for problems where signals are known to be constructed of some constituents, such as patches of natural scenes and texture images. The challenging task associated with this representation is the learning of a discriminate dictionary using the signal constituents. For some problems, like texture analysis, this task has been extensively studied, and the clustering of many candidate constituents has been considered as one of the effective methods to learn the dictionary. Moreover, some signal decomposition algorithms like non-negative matrix factorization (NNMF) are known to decompose signals into their parts~\cite{DL:Lee99,DL:Biggs08} and can accordingly be used to learn the dictionary in this case.

From the \emph{learning} perspective, the dictionary $\mathbf{D}$ is learned:

\paragraph*{i) per class} (Subsection~\ref{subsec:perClass}) While the methods in this category are more simple and computationally efficient, they suffer from two main shortcomings; (1) there might be a redundancy between the atoms in the learned dictionaries of different classes (e.g., a signal constituent that is common to more than one class), and (2) the methods can easily ignore very descriptive atoms that are functions of data instances from different classes (e.g., a metaface that combines positive features from one class and negative features from the other).

\paragraph*{ii) unsupervised learning with supervised pruning} (Subsection~\ref{subsec:Pruning}) In this approach, a large dictionary is first learned unsupervised. The class labels are only included in the pruning step.
The initial large dictionary and subsequent pruning step, however, increases the computational complexity. Moreover, the ultimate discrimination power of the pruned dictionary is always less than the initial large one and therefore, the highest classification performance depends on the discrimination power of the initial large dictionary.

\paragraph*{iii) using all class labels} (Subsections~\ref{subsec:jointDLClassifier} and \ref{subsec:CatInfoDL}) This category of methods solves a usually complex optimization problem that maximizes the descriptiveness of the atoms while minimizing their redundancy. Most of methods are however computationally demanding as the optimization problem has to be solved iteratively and alternately among the dictionary, coefficients, and even classifier parameters.


\subsection{The coefficients $\mathbf{A}$}
Each data instance $\mathbf{x}_{i}$ can be represented in terms of dictionary atoms using the coefficient vector $\bm{\alpha}_i$. These coefficients are usually learned such that the reconstruction error of the original data instance (or its parts) using the dictionary atoms is minimized as can be seen from~(\ref{eq:lasso}).

From the \emph{representation} perspective, a coefficient vector for a data instance represents:
\paragraph*{i) a linear combination of atoms} This is the most common representation of the coefficient vector. Given a data instance and the dictionary atoms, the data instance is usually represented as a sparse combination of dictionary atoms. This representation is suitable to the cases where the dictionary atoms can be used to reconstruct the original data instance.

\paragraph*{ii) a histogram over atoms} This representation is used when  the dictionary atoms represent constituents of the data instances (Subsection~\ref{subsec:histogram}). In this case, the constituents of the new data instance are first selected, and then the coefficient vector is represented as a histogram over the closest atoms for these constituents in the dictionary.

From the \emph{learning} perspective, the coefficients $\mathbf{A}$ are learned:

\paragraph*{i) for test samples only} Some of the S-DLSR algorithms do not require the learning of coefficients for the training instances. Instead, the training data are used for constructing (or learning) the dictionary and then the coefficients are only learned for the new test samples. These methods usually use a simple classification model over the learned dictionary atoms (like nearest neighbour classifiers or the minimum residual classifier). Examples of these algorithms are sparse representation-based classification (SRC)~\cite{DL:Wright09}.

\paragraph*{ii) for training and test samples}  When a complex classification model (like SVM) needs to be learned, the coefficients matrix corresponding to training samples $\mathbf{A}\in\mathbb{R}^{k\times n}$ are also needed. These coefficients represent training instances in the space of dictionary atoms. 
The learning of coefficients can be done separately after the dictionary is learned, when the dictionary has closed-form solution, such as in the HSIC-based S-DLSR~\cite{Gangeh:TSP13}, or simultaneously with the dictionary, which is the case in most S-DLSR methods, where there is no closed-form solution for the dictionary. Moreover, the category information can be used to learn more discriminative coefficients (see Subsection~\ref{subsec:CatInfoCoeff}).

\subsection{The classification model $\mathbf{W}$}

In S-DLSR methods, the classifier receives as an input the encoding of a new data instance in the space of dictionary atoms and returns the encoding of the data instance in the space of classes.

From the \emph{representation} perspective, the classifier can be:
\paragraph*{i) a binary map from atoms to classes} This is the simplest representation used by the S-DLSR methods, in which each group of dictionary atoms maps to a separate class. A new data instance is first mapped to the space of atoms and then a simple classification rule is employed to assign this new instance to one of the classes. This approach is computationally efficient and it is easy to interpret the classification decisions by inspecting the atoms of the assigned class. However, this simple classification rule cannot handle complex class assignment where data points are not directly mapped to the atoms of a single class.

\paragraph*{ii) a linear map from atoms to classes} For linear classifiers, the classifier parameters $\mathbf{W}$ form a mapping from the space of dictionary atoms to that of the classes. In this case, the coefficients matrix $\mathbf{A}$ needs to be learned for the training data and then a linear classification model is learned over these coefficients. In some algorithms, the learning of classifier is done simultaneously with the learning of dictionary (see Subsection~\ref{subsec:jointDLClassifier}). This approach is more complex than the first one but usually results in better classification decisions.

\paragraph*{iii) a non-linear map from atoms to classes} When the data instances in the space of atoms are not linearly separable, one might consider learning a nonlinear classifier (such as SVM with an RBF kernel) over the coefficients matrix $\mathbf{A}$. The use of nonlinear classifiers, however, makes it more computationally difficult to simultaneously learn the dictionary and/or coefficients with the classification models.

From the \emph{learning} perspective, a classification model is learned:
\paragraph*{i) separately, after learning $\mathbf{D}$ and $\mathbf{A}$} Given the representation of the data in the space of atoms, traditional algorithm for supervised learning can be used for learning a classification model for the problem at hand. These methods are more simple and they allow the different existing algorithms to be used with the learned dictionary. One the other hand, learning the coefficient in isolation from the classification model might result in a representation of the data instances that does not necessarily capture the separation between different classes.
\paragraph*{ii) while learning $\mathbf{D}$ and/or $\mathbf{A}$} This approach is more computationally demanding than the first category but it allows for the learning of a representation in which data instances from different classes are well separated in the space of dictionary atoms. This can potentially result in better classification decisions. However, the joint optimization problem obtained for learning both the dictionary and classifier parameters is non-convex, which has to be solved iteratively and alternately. The non-convex optimization problem may lead to some local minima, i.e., sub-optimal solutions. Moreover, due to the complexity of the joint optimization problem, linear classifiers are mainly used, and they may not preform adequately well in more subtle classification tasks~\cite{DL:Mairal08a}.

Tables~\ref{tab:rep} and~\ref{tab:learn} provide a summary of the discussion provided in this section for the three building blocks of an S-DLSR method from representation and learning perspectives, respectively.

\begin{table*}[!t]
  \tiny
  \caption{The representation of different components of an S-DLSR solution.}
  \centering
  \label{tab:rep}
  \begin{tabular}{|p{2cm}|p{2.8cm}|p{7cm}|}
\hline
Component & Representation & Summary \tabularnewline
\hline
\hline
\multirow{4}{3cm}{\textbf{Dictionary $\mathbf{D}$ }} &
\multicolumn{2}{l|}{Each atom can be represented as:}\tabularnewline
\cline{2-3}
 &  a single data instance &
 - easy to interpret, efficient if training data is small

 - sensitive to noisy training instances, inefficient and infeasible
to store and transfer if training data is large
\tabularnewline
\cline{2-3}
 & a function of multiple data instances &
 - smaller dictionary, simple learning algorithms

 - more difficult to interpret, size increases with the number of classes
\tabularnewline
\cline{2-3}
 & a function of signal constituents &
 - smaller dictionary, suitable when signals are constructed of some
constituents

 - more difficult to interpret
\tabularnewline
\hline
\hline
\multirow{3}{3cm}{\textbf{Coefficients $\mathbf{A}$}} &
\multicolumn{2}{l|}{A coefficient vector for a data instance can be represented as:}
\tabularnewline
\cline{2-3}
 & a linear combination of atoms & suitable when atoms can reconstruct the signals
\tabularnewline
\cline{2-3}
 & a histogram over atoms & suitable when atoms represent constituents of signals
\tabularnewline
\hline
\hline
\multirow{4}{3cm}{\textbf{Classifier $\mathbf{W}$}} &
\multicolumn{2}{l|}{A classification model can be represented as:}\tabularnewline
\cline{2-3}
 & a binary map from atoms to classes &
 - computationally efficient and easy to interpret

 - cannot handle complex class assignment
\tabularnewline
\cline{2-3}
 & a linear map from atoms to classes &
 - more accurate

 - more computationally complex
\tabularnewline
\cline{2-3}
 & a non-linear map from atoms to classes &
 - suitable for complex data with non-linear classes

 - computationally infeasible to learn simultaneously with \textbf{$\mathbf{D}$}
and $\mathbf{A}$
\tabularnewline
\hline
\end{tabular}

\end{table*}

\begin{table*}[!t]
  \tiny
  \caption{The learning of different components of an S-DLSR solution.}
  \centering
  \label{tab:learn}

  \begin{tabular}{|p{2cm}|p{2.8cm}|p{7cm}|}
\hline
Component & Learning & Summary \tabularnewline
\hline
\hline
\multirow{3}{3cm}{\textbf{Dictionary $\mathbf{D}$}} & per class &
- simple and computationally efficient

- redundancy among atoms, prone to ignoring descriptive atoms
\tabularnewline
\cline{2-3}
 & unsupervised learning with supervised pruning &
 - less redundancy among atoms

 - computationally complex
\tabularnewline
\cline{2-3}
 & using all class labels &
 - more optimal in terms of redundancy and descriptiveness

 - complex optimization, computationally demanding
\tabularnewline
\hline
\hline
\multirow{2}{3cm}{\textbf{Coefficients $\mathbf{A}$}} & for test samples only &
- simple classification model

- less accurate
\tabularnewline
\cline{2-3}
 & for training and test samples &
 - more accurate for complex data

 - more computationally demanding
\tabularnewline
\hline
\hline
\multirow{2}{3cm}{\textbf{Classifier $\mathbf{W}$}} & separately, after learning $\mathbf{D}$ and $\mathbf{A}$ &
- simpler, usable with different DL algorithms

- less separation between classes\tabularnewline
\cline{2-3}
 & while learning $\mathbf{D}$ and $\mathbf{A}$ &
- more classification accuracy for linear classifiers

- computationally complex, sub-optimal solutions
\tabularnewline
\hline
\end{tabular}

\end{table*}

\section{Conclusion}\label{sec:Conclusion}


Supervised dictionary learning and sparse representation (S-DLSR) is an emerging category of methods that result in more optimal dictionary and sparse representation in classification tasks. In this paper, we surveyed the state-of-the-art techniques for S-DLSR and presented a comprehensive views of these techniques. We have identified six main categories of S-DLSR methods and highlighted the advantages and shortcomings of the methods in each category. Furthermore, we have provided a summary of the building blocks for an S-DLSR method including the dictionary, sparse coefficients, and classifier parameters from two perspectives: representation and learning. This enables the researchers to decide on how to choose these blocks to design a new S-DLSR algorithm based on the problem at hand. This review addresses a gap in the literature and is anticipated to advance the research in S-DLSR and its applicability to a variety of domains.

\section*{Acknowledgements}

This work was supported in part by the Natural Sciences and Engineering Research Council (NSERC) of Canada under Canada Graduate Scholarship (CGS D3-378361-2009), and in part by the NSERC Postdoctoral Fellowship (PDF-454649-2014).





\bibliographystyle{model1-num-names}
\bibliography{D:/Mehrdad/Bib_Centeralized/refNCD,D:/Mehrdad/Bib_Centeralized/refTexture}







\end{document}